\pdfoutput=1
\documentclass[a4paper,english,12pt]{article}
\usepackage[utf8]{inputenc}
\usepackage[T2A]{fontenc}
\usepackage{graphicx}%
\usepackage{multirow}%
\usepackage{amsmath,amssymb,amsfonts}%
\usepackage{amsthm}%
\usepackage{mathrsfs}%
\usepackage[title]{appendix}%
\usepackage{xcolor}%
\usepackage{textcomp}%
\usepackage{manyfoot}%
\usepackage{booktabs}%
\usepackage{listings}%
\usepackage{adjustbox}
\usepackage{hyperref,booktabs,cite,enumerate,amsmath,amssymb}
\usepackage{paralist,tikz,pgfplots,multirow,lscape,multicol,longtable}
\usepackage{rotating,libertine}
\usepackage{authblk}

\usepackage{fvextra}
\usepackage{newfloat}
\DeclareFloatingEnvironment[name=Listing,placement=htbp]{listing}

\usepackage[right=20mm,left=25mm,top=20mm,bottom=20mm]{geometry}

\usepackage[main=english,russian]{babel}
\usepackage[table]{xcolor}
\usepackage{paralist,caption}

\newcommand{\kyr}[1]{{\foreignlanguage{russian}{#1%
}}}

\pgfplotsset{compat=1.18}

\begin{document}

\title{KyrgyzLLM-Bench: Benchmarking \\Kyrgyz Language Understanding}

\renewcommand\Authsep{\qquad}
\renewcommand\Authand{\qquad}
\renewcommand\Authands{\qquad}
\setlength\affilsep{0.4em}

\author[1]{Timur Turatali\thanks{\href{mailto:timur.turat@gmail.com}{timur.turat@gmail.com}}}
\author[2]{Aida Turdubaeva}
\author[3]{Rustem Izmailov}
\author[2,4,5]{Anton~M.~Alekseev}
\author[4,5]{Sergey~I.~Nikolenko}

\affil[1]{The Cramer Project}
\affil[2]{Institute of IT, Kyrgyz State Technical University named after I.~Razzakov}
\affil[3]{School of Computer Science, University of Windsor}
\affil[4]{St.~Petersburg Department of the Steklov Math. Institute, RAS}
\affil[5]{St.~Petersburg State University}

\date{}
\maketitle

\vspace{-1.2cm}

\begin{abstract}
Evaluating large language models (LLMs) across languages remains challenging, as most multilingual benchmarks rely on translated English datasets, often obscuring linguistic and cultural specificity in the target language. This issue is particularly pronounced for less-resourced languages such as Kyrgyz, where reliable natively authored evaluation data are scarce. Building on previously introduced Kyrgyz-language evaluation datasets, this work reports the first systematic and large-scale evaluation of LLMs in Kyrgyz using the KyrgyzLLM-Bench benchmark suite. KyrgyzLLM-Bench comprises two natively authored datasets---\textit{KyrgyzMMLU} and \textit{KyrgyzRC}---together with carefully translated and manually post-edited versions of \textit{WinoGrande}, \textit{HellaSwag}, \textit{BoolQ}, and \textit{TruthfulQA}. We evaluate 26 open- and closed-source LLMs under zero-shot and few-shot settings, analyzing model performance, cross-lingual transfer, and the impact of translation artifacts on evaluation reliability. {Across families and tasks, model rankings transfer broadly from English to Kyrgyz on \textit{WinoGrande}} {and \textit{BoolQ}, and to a lesser extent on \textit{MMLU}}{, while \textit{HellaSwag} exhibits a substantial English--Kyrgyz performance gap consistent with translation-induced plausibility shifts. Few-shot prompting improves several open-source models on reading comprehension but behaves inconsistently for proprietary models on translated tasks. We publicly release all datasets, evaluation code, and per-model results, and integrate the Kyrgyz tasks into a widely used multilingual evaluation framework to support future research on Kyrgyz NLP.\footnote{Preprint. This version has not undergone peer review.}}


\end{abstract}

\section{Introduction}

The rapid progress of large language models (LLMs) has increased the need for robust and diverse evaluation benchmarks.
Suites such as MMLU~\cite{hendrycks2021measuring} have become a de~facto standard for assessing reasoning and knowledge capabilities.
However, current evaluations remain heavily skewed toward English~\cite{wu2025bitter}, leaving substantial gaps in our understanding of model performance across diverse linguistic and cultural contexts.

To broaden multilingual evaluation, a number of benchmarks have been developed, most commonly by machine-translating English datasets~\cite{lai2023okapi}.
While pragmatic, this approach introduces well-documented issues, including subtle translation errors, unnatural phrasing, and ``translationese''~\cite{vanmassenhove2021}.
More critically, translated benchmarks often retain cultural assumptions specific to the source material rather than reflecting the cultural and contextual grounding of the target-language community, with measurable effects on evaluation reliability in culture-sensitive domains such as social sciences, history, and literature~\cite{singh-etal-2025-global}.
For less-resourced languages such as Kyrgyz, the lack of high-quality, natively curated evaluation data continues to hinder both systematic research and the development of reliable language technologies.

In this work, we report a systematic and large-scale evaluation of LLMs in Kyrgyz using \emph{KyrgyzLLM-Bench}, a multi-faceted benchmark suite based on previously introduced Kyrgyz-language evaluation datasets.
In contrast to evaluations relying solely on translated data, the core components of KyrgyzLLM-Bench are natively authored in Kyrgyz, ensuring linguistic naturalness and cultural relevance.

KyrgyzLLM-Bench comprises:
\begin{inparaenum}[(1)]
    \item \textit{KyrgyzMMLU}, a large-scale multitask multiple-choice question-answering benchmark with $7{,}977$ items written by curriculum experts and aligned with the Kyrgyz national education registry, covering subjects such as mathematics, physics, literature, and history;
    \item \textit{KyrgyzRC}, a native reading-comprehension dataset consisting of $400$ questions based on authentic Kyrgyz texts, requiring contextual understanding and multi-sentence reasoning;
    \item translated benchmarks: manually post-edited Kyrgyz versions of \textit{WinoGrande}, \textit{HellaSwag}, \textit{BoolQ}, and \textit{TruthfulQA}, ensuring linguistic fidelity and cultural appropriateness.
\end{inparaenum}

This paper substantially extends an earlier conference paper~\cite{turatali2025bridging}, presented at TurkLang-2025, which introduced the \textit{KyrgyzLLM-Bench} resources to the Kyrgyz NLP community. Relative to that work, the present journal version provides a more detailed account of dataset construction, annotation, and quality-control protocols; an expanded evaluation covering 26 open- and closed-source LLMs under zero-shot and few-shot regimes---providing, to our knowledge, the first systematic analysis of LLM performance on complex, culturally grounded Kyrgyz tasks at this scale---complemented by parallel English-language baselines that support the cross-lingual analyses in Sections~\ref{sec:evaluation-results} and~\ref{sec:discussion}; integration of the Kyrgyz tasks into the \emph{Lighteval} evaluation framework; and new analyses of language-specific characteristics relevant to evaluation (Section~\ref{sec:kyrgyz}), translation-induced plausibility shifts, and few-shot stability. We publicly release all datasets, evaluation code, and per-model results to support future research on Kyrgyz NLP.

Based on prior work on translated benchmarks and cross-lingual evaluation, we put forward the following hypotheses:
\begin{inparaenum}[(1)]
     \item core reasoning and question-answering capabilities partially transfer from English to Kyrgyz, preserving relative model rankings on structurally similar tasks such as \textit{BoolQ} and \textit{WinoGrande};
    \item event-continuation benchmarks that rely on plausibility judgments, such as \textit{HellaSwag}, are particularly sensitive to translation artifacts, leading to \emph{plausibility shifts}{---unpredictable changes in perceived naturalness and coherence that compromise the reliability of cross-lingual measurements.}
\end{inparaenum}

In the remainder of the paper, we first review the relevant prior research and describe the construction of KyrgyzLLM-Bench, then present extensive evaluations of open-source and proprietary LLMs, and finally discuss cross-lingual transfer, robustness, and implications for low-resource evaluation.

\section{Related Work}

Large language models (LLMs) have achieved strong performance on a wide range of natural language understanding and reasoning benchmarks, including \textit{GLUE}~\cite{wang2018glue}, \textit{SuperGLUE}~\cite{wang2019superglue}, and \textit{MMLU}~\cite{hendrycks2021measuring}, as well as commonsense reasoning datasets such as \textit{WinoGrande}~\cite{sakaguchi2020winogrande_ieee}, \textit{HellaSwag}~\cite{zellers2019hellaswag_ieee}, and \textit{GSM8K}~\cite{cobbe2021training_ieee}.
However, these benchmarks are primarily available in English and other high-resource languages, with limited coverage for low-resource languages such as Kyrgyz.
To address this gap, multilingual benchmarks such as \textit{XTREME}~\cite{hu2020xtreme} and \textit{FLORES}~\cite{goyal2022flores} have been developed to evaluate cross-lingual transfer across typologically diverse languages.
Yet many low-resource languages, particularly Central Asian and Turkic languages such as Kyrgyz, remain excluded.
One practical approach to creating benchmarks for low-resource languages is to translate existing English datasets.
For example, \textit{MMLU} and \textit{COPA} have been translated into Latvian~\cite{skadina2025evaluation}, although these translations often lack manual post-editing and may introduce additional noise.
The \textit{OKAPI} framework~\cite{lai2023okapi} translates ARC, HellaSwag, and MMLU into 26 languages using ChatGPT, but its language coverage does not include Kyrgyz or any other Turkic language, and translations are not validated with native speakers.
In this study, four widely used benchmarks---\textit{WinoGrande}, \textit{HellaSwag}, \textit{BoolQ}, and \textit{TruthfulQA}---are translated into Kyrgyz and manually reviewed to ensure cultural and linguistic appropriateness.
In addition, the benchmark includes original Kyrgyz-language tasks, including a localized \textit{MMLU} and reading comprehension tests based on official Kyrgyz school exams, whose construction and annotation procedures are described in detail in this paper.
This approach is consistent with that of~\cite{dargis2024evaluating}, who used centralized Latvian high school exams to evaluate LLM performance.

By combining translated datasets with culturally grounded, original benchmarks, this work provides new insights into multilingual generalization and the capabilities of open-source LLMs for Kyrgyz, a Turkic language spoken by approximately four million people (about 80\% of the population) in the Kyrgyz Republic~\cite{Salmorbekova2023KyrgyzRussian}.

{To our knowledge, no manually validated MMLU-style benchmark and no native reading-comprehension benchmark for Kyrgyz existed prior to this study, and the centralized-examination evaluation logic of~\cite{dargis2024evaluating} is, to our knowledge, applied here to a Turkic less-resourced language for the first time. We position KyrgyzLLM-Bench as complementing rather than replacing prior translation-based efforts: it adds the natively authored, expert-validated component that was missing.}

{\section{Kyrgyz Language Characteristics Relevant to~Evaluation}\label{sec:kyrgyz}

This section briefly outlines the properties of Kyrgyz that are most relevant to LLM evaluation, rather than providing a comprehensive linguistic description; for the latter, we refer the reader to~\cite{alekseev2024kyrgyznlp}. Three properties of Kyrgyz are particularly relevant for benchmarking and translation-based evaluation.

\textbf{Agglutinative Turkic morphology.} Kyrgyz is an agglutinative Turkic language in which grammatical relations and derivational distinctions are expressed primarily through productive suffix concatenation~\cite{alekseev2024kyrgyznlp}. As a consequence, surface word forms are typically longer than corresponding English forms and far less likely to appear as single tokens in subword vocabularies trained predominantly on Indo-European data. This affects evaluation in two practical ways: tokenizer fragmentation increases effective context length for the same semantic content, and lexical generalization is weakened when morphologically related forms are split into rare or unrelated subword sequences.

\textbf{Cyrillic script with extended graphemes.} Kyrgyz uses a Cyrillic-based alphabet that includes letters absent from standard Russian (notably \kyr{ң}, \kyr{ү}, \kyr{ө}). Tokenizers built without dedicated Kyrgyz coverage may map these graphemes to byte-level fallback or treat them as low-frequency tokens, which influences both input encoding and the parsing of generated answer strings.

\textbf{Limited representation in pretraining and resource scarcity.} Kyrgyz remains substantially underrepresented in mainstream pretraining corpora and in publicly available NLP resources~\cite{alekseev2024kyrgyznlp,mirzakhalov2021evaluating,veitsman2025}. Despite recent commercial and non-commercial efforts~\cite{akylai_24kg,sexed_kyrgyz}, manually annotated datasets for core language-processing tasks remain scarce.

These properties bear on the patterns we observe in Section~\ref{sec:evaluation-results}. The pronounced English--Kyrgyz gap on \textit{HellaSwag} is consistent with translation-induced disruption of agglutinative event-completion phrasing, where idiomatic continuations rely on morphological cohesion that does not survive literal translation. The relatively low and unstable accuracy on \textit{KyrgyzMMLU} subjects with strong language-specific content (\textit{Kyrgyz Language}, \textit{Kyrgyz Literature}; see Tables~\ref{tab:ky_mmlu_zs_and_delta} and~\ref{tab:kg_combined_results}) compared with cross-linguistically transferable subjects such as \textit{Mathematics} or \textit{Physics} is consistent with limited Kyrgyz-specific knowledge in pretraining. We do not claim a causal attribution to any single factor; we point to these features as the most plausible explanatory candidates supported by the existing Kyrgyz NLP literature.}

\section{KyrgyzLLM-Bench Construction}\label{sec:benchmark}

{\paragraph{Comparison with similar resources.} KyrgyzLLM-Bench differs from prior multilingual evaluation efforts along} {two axes.} {\emph{Authorship and validation}: \textit{KyrgyzMMLU} and \textit{KyrgyzRC} are natively authored from official Kyrgyz educational materials and authentic Kyrgyz texts, with each item reviewed by domain experts and Kyrgyz language educators. This mitigates the kinds of translation-noise issues documented in recent quality analyses of automatically-curated Kyrgyz language resources; for instance, an independent qualitative analysis of the public OPUS sample of the NLLB v1 English--Kyrgyz parallel corpus reported that only about a third of sample sentence pairs were high-quality, usable translations, with the remainder exhibiting language-identification errors, misalignment, or fluency problems~\cite{jumashev2025kyrgyz}. \emph{Scope}: \textit{KyrgyzMMLU} provides $7{,}977$ items spanning} {eight school subjects plus a Medicine specialization,} {and \textit{KyrgyzRC} adds a $400$-item native reading-comprehension component covering encyclopedic, journalistic, literary, and mathematical genres; we are not aware of comparable Kyrgyz-language coverage in any prior public benchmark.}

\subsection{Original benchmark: KyrgyzMMLU}

\textit{KyrgyzMMLU} is based on materials from the General Republican Testing (GRT), conducted in Kyrgyzstan since 2002.
The GRT is administered by the Center for Educational Assessment and Teaching Methods (CEATM) in cooperation with the Ministry of Education of the Kyrgyz Republic.
The test set was officially sourced from the Department for Development of Education Quality under the Ministry of Education and Science and comprises $8$ school subjects taught from 6th to 11th grade, as well as specialized categories such as Medicine.
The subject distribution is summarized in Table~\ref{tab:subjects}.
The GRT aims to ensure equal access to higher education through fair and independent testing.
The test assesses applicants' ability to successfully continue their studies at a higher education institution and is conducted in Kyrgyz and Russian.
The assessment targets reasoning skills and the application of school knowledge.

\begin{table}[t]
\centering
\begin{minipage}[t]{0.48\linewidth}
\centering\footnotesize
    \begin{tabular}{p{.8\linewidth}r}
    \toprule
    \textbf{School subject} & \textbf{\#Q} \\
    \midrule
    Mathematics              & 1{,}169 \\
    Biology (Bio)            & 1{,}550 \\
    Physics (Phys)           & 1{,}228 \\
    Chemistry (Chem)         & 1{,}205 \\
    Kyrgyz Literature (Lit)  & 1{,}169 \\
    Geography (Geog)         &   640 \\
    Kyrgyz History (Hist)    &   440 \\
    Kyrgyz Language (Lang)   &   360 \\
    Medicine (Med)           &   216 \\
    \bottomrule
    \end{tabular}
\captionof{table}{School subjects in~\textit{KyrgyzMMLU}.}
\label{tab:subjects}
\end{minipage}$\qquad$
\begin{minipage}[t]{0.4\linewidth}
\centering\footnotesize
    \begin{tabular}{p{.8\linewidth}r}
    \toprule
    \textbf{Subject} & \textbf{\#Q} \\
    \midrule
    Math              & 100 \\
    Kyrgyz Wikipedia  & 100 \\
    Kyrgyz News       & 100 \\
    Kyrgyz Literature & 100 \\
    \bottomrule
    \end{tabular}
\captionof{table}{Subjects in~KyrgyzRC.}
\label{tab:kyrgyzrcsubjects}
\end{minipage}
\end{table}

\begin{table}[t]
\centering
\caption{Sample test task: ``If the post office takes 10\% of the total amount for a money transfer, then how many {extra} soms will Askar pay to send 600 soms?'' The correct answer is \kyr{(В)}~60.}
\label{tab:kymmlu-example}
\small
    \begin{tabular}{@{}p{\linewidth}@{}}
    \toprule
    \kyr{\textbf{Question:} Эгерде почта аркылуу акча жиберүүнүн кызматы үчүн жиберилүүчү сумманын 10\% ын төлөө керек болсо, анда Аскар 600 сомду жиберүү үчүн канча сом төлөйт?} \\[4pt]
    \kyr{\textbf{Options:} (А)~6 \quad (Б)~10 \quad (В)~60 \quad (Г)~100 \quad (Д)~160} \\
    \bottomrule
    \end{tabular}
\end{table}

\begin{table}[t]
\centering
\caption{A reading comprehension task: ``The Great Kyrgyz Khaganate is the name of the Yenisei Kyrgyz state during its height of power in the IX century. In 840, after defeating the Uyghur Khaganate, the Great Kyrgyz state occupied territories from the Orkhon to East Turkestan, and from the Sayan-Altai to the Syr Darya. This era in Kyrgyz history is called the Kyrgyz Great Power. The Khaganate existed until 924. \textbf{Q}: In which century was the Great Kyrgyz Khaganate at its peak?'' (\textbf\kyr{(А)} 9-кылымда, in the IX century).}
\label{tab:kyrgyzrc-example}
\small
    \begin{tabular}{@{}p{\linewidth}@{}}
    \toprule
    \kyr{ \textbf{Text:} Улуу Кыргыз кагандыгы -- 9-кылымда Енисей Кыргыз мамлекетинин күчөп турган мезгилиндеги расмий аталышы. 840-жылы Уйгур кагандыгын талкалап, Улуу Кыргыз дөөлөтү Орхондон Чыгыш Түркстанга, Саян-Алтайдан Сыр-Дарыяга чейинки аймактарды ээлеген. Бул доор кыргыз тарыхында Кыргыз улуу державасы деп аталган. Кагандык 924-жылга чейин жашаган.} \\[4pt]
    \kyr{\textbf{Question:} Улуу Кыргыз кагандыгы кайсы кылымда күчөп турган?} \\[4pt]
    \kyr{\textbf{Options:} (А)~9-кылымда. \quad (Б)~8-кылымда. \quad (В)~10-кылымда. \quad (Г)~7-кылымда.} \\
    \bottomrule
    \end{tabular}
\end{table}

{The choice against tests focused exclusively on school subjects reflects structural realities in Kyrgyzstani education: while all applicants nationwide must be assessed by a single instrument, teaching quality varies considerably across regions due to shortages of qualified teachers (mainly in rural areas), gaps in educational materials, and uneven access to technical resources and mass media. Applicants therefore enter the assessment with markedly different levels of school-acquired knowledge and skills. The GRT was designed to mitigate these disparities, motivating the choice of a common test structure taken by all applicants.}

The Kyrgyzstani GRT uses a multiple-choice format with one correct option ({Table~\ref{tab:kymmlu-example}} shows a sample question). The primary metric for \emph{KyrgyzMMLU} and \emph{KyrgyzRC} is accuracy.

\subsection{Original benchmark: KyrgyzRC}\label{subsec:kyrgyzrc}

\textit{KyrgyzRC} is a natively authored reading-comprehension dataset designed to evaluate understanding and reasoning in Kyrgyz.
It consists of $400$ manually curated multiple-choice questions drawn from diverse sources, including Kyrgyz Wikipedia, national news articles, literary excerpts, and school-level math problems {(Table~\ref{tab:kyrgyzrcsubjects})}.
Each item consists of a 2--5 sentence passage followed by a question and four answer options, with exactly one correct answer ({Table~\ref{tab:kyrgyzrc-example}}).

{\paragraph{Authorship and construction workflow.} \textit{KyrgyzRC} was
authored by $19$ students from the Department of Computational Linguistics at
the Kyrgyz State Technical University named after I.~Razzakov, all native Kyrgyz
speakers, as part of their supervised coursework (see the Statements and
Declarations for participant consent details). Each of the four source
domains---Kyrgyz Wikipedia, national news, Kyrgyz literature, and school-level
mathematics---contributes $100$ items, for a total of $400$ questions. The
annotation pipeline proceeded in three sequential stages:}
\begin{inparaenum}[(1)]
    \item {\emph{Item creation.} Each student was assigned passages from
    a specific domain and authored a question with four answer
    options---one correct and three plausible distractors---targeting one of
    the four comprehension-skill categories listed above. Authors were
    instructed to keep passages between two and five sentences, to design
    distractors that were topically related to the passage but unambiguously
    incorrect, and to avoid items with more than one defensible answer.}
    \item {\emph{Domain-level curation.} Four supervisors, one per
    domain, reviewed all items within their assigned topic for linguistic
    accuracy, answer correctness, distractor plausibility, and consistency
    with the intended comprehension-skill label.}
    \item {\emph{Final linguistic review.} A dedicated professional
    linguist conducted a pass over the entire dataset, checking for uniform
    formatting, natural phrasing in Kyrgyz, and the absence of ambiguous or
    doubly-correct items. Items flagged at this stage were either revised in
    consultation with the original author or replaced.}
\end{inparaenum}

{The same cohort of $19$ students and $4$ supervisors subsequently
undertook the translated-benchmarks post-editing workflow described in
Section~\ref{subsec:translated}; the participant counts cited there refer to
the same individuals, not a separate group.}

{\paragraph{Metadata schema.} Each \textit{KyrgyzRC} item is annotated with four metadata fields: (i)~\emph{source type} (Kyrgyz Wikipedia, national news, Kyrgyz literature, or school-level mathematics); (ii)~\emph{question type} (factual understanding, inference, vocabulary in context, or reasoning across sentences); (iii)~the index of the correct answer; and (iv)~the passage text from which the question was derived. A sample data entry following this schema is provided in Appendix~\ref{appsec:kyrgyzrc-entry}.}

{\textit{KyrgyzRC} adopts a multiple-choice format to enable automatic evaluation and consistent scoring, mirroring standardized assessments used in Kyrgyz education. The balance across encyclopedic, journalistic, literary, and mathematical genres supports evaluation across varied linguistic registers and domains. Each entry includes metadata for source type and question type.}

{\textit{KyrgyzRC} is, to our knowledge, the first publicly available reading-comprehension benchmark designed specifically for Kyrgyz. It addresses a key gap in evaluating context-sensitive understanding for a less-resourced language. \textit{KyrgyzRC} is publicly released and can serve both as a benchmark and as a resource for developing and testing Kyrgyz language models.}

\subsection{Translated benchmarks}\label{subsec:translated}

To complement the natively authored datasets, four widely used English benchmarks were translated into Kyrgyz.
Specifically:
\begin{inparaenum}[(1)]
    \item commonsense reasoning---\textit{HellaSwag}~\cite{zellers2019hellaswag_ieee}, which tests plausible sentence continuation, and \textit{WinoGrande}~\cite{sakaguchi2020winogrande_ieee}, which tests pronoun resolution in context;
    \item reading comprehension---\textit{BoolQ}~\cite{clark2019boolq}, which requires answering natural-language questions given a short context;
    \item robustness and factuality---\textit{TruthfulQA}~\cite{lin2022truthfulqa}, which probes a model's tendency to produce truthful answers rather than repeat common misconceptions.
\end{inparaenum}
\textit{GSM8K}~\cite{cobbe2021training_ieee} was also translated into Kyrgyz; however, it was excluded from this study for the reasons described in Appendix~\ref{appsec:gsm8k}.

Collectively, this set provides a compact, high-signal evaluation of core LLM understanding, reasoning, and robustness. All tasks also appear in the \emph{Lighteval} tool and fall into its higher-level benchmark categories.\footnote{See \emph{Lighteval}'s README: \url{https://github.com/huggingface/lighteval/blob/main/README.md}}
{Table~\ref{tab:kywinogrande-example}} shows an example from the translated \textit{WinoGrande} benchmark.

\begin{table}[t]
\centering
\caption{A translated \textit{WinoGrande} data point.}
\label{tab:kywinogrande-example}
\small
    \begin{tabular}{@{}p{\linewidth}@{}}
    \toprule
    \textbf{Original English:} He never comes to my home, but I always go to his house because the \_ is smaller. \\[2pt]
    \textbf{Options:} (1)~home \quad (2)~house \\[4pt]
    \midrule
    \kyr{\textbf{Translated Kyrgyz:} Ал менин үйүмө эч качан келбейт, бирок мен ар дайым анын турак жайына барам, анткени \_ кичинекей.} \\[2pt]
    \kyr{\textbf{Options:} (1)~үй \quad (2)~турак жай} \\
    \bottomrule
    \end{tabular}
\end{table}

For each dataset, the following quality-control procedures were applied:
\begin{inparaenum}[(1)]
    \item {automatic translation}: source examples were first translated into Kyrgyz by \emph{Claude 4 Sonnet} and independently by \emph{{Gemini 2.5 Flash}};
    \item ensemble validation: the two outputs were compared to identify lexical or semantic divergences;
    \item manual post-editing: Kyrgyz linguists and domain experts reviewed all examples to resolve ambiguities, preserve idiomatic usage, and ensure cultural appropriateness;
    \item {quality assurance}: back-translation checks and spot-checks of 10\% of entries confirmed fidelity to original meaning.
\end{inparaenum}

The translation and validation were conducted in an academic setting as part of a supervised university course at the Department of Computational Linguistics.
A total of $19$ students and $4$ supervisors/curators, all native Kyrgyz speakers, participated in the process.
The workflow followed a peer-review structure: each instance was edited by one student and independently verified by another, with oversight from course instructors. {Details on participant consent and data-handling are provided in the Statements and Declarations section.}

\begin{table}[htbp]
\centering
\resizebox{\linewidth}{!}{%
\setlength{\tabcolsep}{3pt}
    \begin{tabular}{l|ccccc|c|ccccc|c}
    \toprule
    & \multicolumn{6}{c|}{\textbf{Zero-shot}}
    & \multicolumn{6}{c}{\textbf{Few-shot} (5-shot; 10-shot for HellaSwag)} \\
    \cmidrule(lr){2-7} \cmidrule(lr){8-13}
    \textbf{Model}
    & \begin{sideways}\textbf{MMLU}\end{sideways}
    & \begin{sideways}\textbf{Wino\allowbreak Grande}\end{sideways}
    & \begin{sideways}\textbf{BoolQ}\end{sideways}
    & \begin{sideways}\textbf{Hella\allowbreak Swag}\end{sideways}
    & \begin{sideways}\textbf{Truthful\allowbreak QA}\end{sideways}
    & \textbf{Avg}
    & \begin{sideways}\textbf{MMLU}\end{sideways}
    & \begin{sideways}\textbf{Wino\allowbreak Grande}\end{sideways}
    & \begin{sideways}\textbf{BoolQ}\end{sideways}
    & \begin{sideways}\textbf{Hella\allowbreak Swag}\end{sideways}
    & \begin{sideways}\textbf{Truthful\allowbreak QA}\end{sideways}
    & \textbf{Avg} \\
    \midrule
    Qwen2.5-0.5B-Instruct & 42.7 & 53.4 & 59.9 & 40.6 & 34.4 & 46.2 & \cellcolor{green!12}43.8 & \cellcolor{green!12}54.5 & \cellcolor{red!10}59.7 & \cellcolor{red!11}39.9 & \cellcolor{green!15}36.7 & \cellcolor{green!12}46.9 \\
    Qwen2.5-1.5B-Instruct & 58.6 & 58.9 & 75.8 & 50.8 & 38.9 & 56.6 & \cellcolor{green!11}59.2 & \cellcolor{green!14}60.7 & \cellcolor{green!16}78.9 & \cellcolor{red!11}50.3 & \cellcolor{green!17}42.6 & \cellcolor{green!15}58.3 \\
    Qwen2.5-3B-Instruct & 64.4 & 66.4 & 68.6 & 56.4 & 50.3 & 61.2 & \cellcolor{green!12}65.3 & \cellcolor{red!12}65.3 & \cellcolor{green!33}79.9 & \cellcolor{red!10}56.3 & \cellcolor{green!14}52.1 & \cellcolor{green!17}63.8 \\
    Qwen2.5-7B-Instruct & 71.5 & 67.1 & 82.5 & \textbf{62.0} & \textbf{56.2} & \textbf{67.9} & \cellcolor{green!15}74.2 & \cellcolor{green!17}70.4 & \cellcolor{green!16}85.4 & \cellcolor{green!12}\textbf{63.2} & \cellcolor{green!12}{\textbf{57.2}} & \cellcolor{green!16}\textbf{70.1} \\
    \midrule
    Qwen3-0.6B & 38.3 & 52.5 & 45.5 & 37.6 & 34.8 & 41.7 & \cellcolor{green!29}47.7 & \cellcolor{green!13}53.8 & \cellcolor{green!25}53.1 & \cellcolor{green!10}37.8 & \cellcolor{green!20}39.6 & \cellcolor{green!23}46.4 \\
    Qwen3-1.7B & 56.0 & 58.3 & 30.3 & 46.1 & 37.6 & 45.7 & \cellcolor{green!22}61.8 & \cellcolor{red!13}56.7 & \cellcolor{green!70}80.3 & \cellcolor{green!10}46.2 & \cellcolor{green!20}42.4 & \cellcolor{green!45}57.5 \\
    Qwen3-4B & 71.6 & 61.8 & 79.6 & 52.3 & 45.8 & 62.2 & \cellcolor{green!12}72.7 & \cellcolor{red!10}61.7 & \cellcolor{green!23}86.0 & \cellcolor{green!17}55.6 & \cellcolor{green!13}47.4 & \cellcolor{green!17}64.7 \\
    Qwen3-8B & \textbf{75.9} & 64.8 & \textbf{85.1} & 57.1 & 45.4 & 65.7 & \cellcolor{green!13}\textbf{77.6} & \cellcolor{green!14}66.9 & \cellcolor{green!15}\textbf{87.6} & \cellcolor{green!12}58.1 & \cellcolor{green!16}48.3 & \cellcolor{green!16}67.7 \\
    \midrule
    Gemma-3-270m & 20.9 & 49.6 & 0.0 & 25.0 & 24.1 & 23.9 & 20.9 & 49.6 & 0.0 & 25.0 & \cellcolor{green!18}28.3 & \cellcolor{green!12}24.8 \\
    Gemma-3-1b-it & 33.8 & 55.6 & 71.4 & 43.4 & 31.5 & 47.1 & \cellcolor{green!16}36.9 & \cellcolor{red!15}53.1 & \cellcolor{red!11}70.9 & \cellcolor{red!12}42.3 & \cellcolor{green!22}37.4 & \cellcolor{green!12}48.1 \\
    Gemma-3-4b-it & 56.5 & 60.7 & 76.6 & 55.9 & 43.3 & 58.6 & \cellcolor{red!11}56.2 & \cellcolor{green!15}63.1 & \cellcolor{red!70}0.0 & \cellcolor{red!70}24.6 & \cellcolor{green!21}48.8 & \cellcolor{red!70}38.5 \\
    \midrule
    Llama-3.2-1B-Instruct & 45.6 & 58.4 & 69.4 & 45.7 & 35.3 & 50.9 & \cellcolor{red!14}43.8 & \cellcolor{red!13}56.7 & \cellcolor{green!13}70.8 & \cellcolor{red!11}45.2 & \cellcolor{green!19}40.0 & \cellcolor{green!11}51.3 \\
    Llama-3.2-3B-Instruct & 58.2 & 64.3 & 72.6 & 53.3 & 42.6 & 58.2 & \cellcolor{green!11}58.6 & \cellcolor{red!10}64.2 & \cellcolor{green!21}78.3 & \cellcolor{red!10}53.2 & \cellcolor{green!19}47.0 & \cellcolor{green!16}60.3 \\
    Llama-3.1-8B-Instruct & 65.0 & \textbf{71.3} & 82.5 & 59.8 & 46.2 & 65.0 & \cellcolor{green!11}65.5 & \cellcolor{green!15}\textbf{74.0} & \cellcolor{green!18}86.7 & \cellcolor{green!14}62.0 & \cellcolor{green!22}52.4 & \cellcolor{green!19}68.1 \\
    \bottomrule
    \end{tabular}%
}
\caption{Zero-shot and few-shot evaluation results on English benchmarks (accuracy \%). Cell colors in the few-shot half show gains/losses compared to the zero-shot case.}
\label{tab:en_combined_results}
\end{table}

\section{Experimental Setup}

We conducted two experimental setups.
The first evaluates $14$ open-source models on the full benchmark, while the second uses a condensed subset (\emph{KyrgyzLLM Tiny Bench}) to evaluate $12$ proprietary models.
We additionally evaluated the original English benchmarks as a cross-lingual reference, selecting {$17$} MMLU subjects analogous to those in KyrgyzMMLU.
\footnote{Selected: {college}\_biology, ...\_chemistry, ...\_mathematics, ...\_medicine, ...\_physics, {high\_school}\_biology, ...\_chemistry, ...\_computer\_science, ...\_european\_history, ...\_geography, ...\_mathematics, ...\_physics, ...\_statistics, ...\_us\_history, ...\_world\_history, prehistory, and professional\_medicine.}
All evaluations were conducted using \emph{Lighteval}~\cite{lighteval}.

{We adopt accuracy with text-based answer parsing as our scoring protocol; this choice keeps the evaluation comparable across open-source and proprietary models, since closed-weight providers expose only generated text via API and do not return token logits suitable for option scoring. Limitations of this choice---in particular sensitivity to output formatting---are discussed in Section~\ref{subsec:limits}.}

{We use temperature $=0.6$ and top-$p=0.9$, a moderately stochastic decoding regime commonly adopted in established LLM inference frameworks and evaluation pipelines~\cite{nemo_inference_guidelines,lm_eval_harness,hf_open_llm_leaderboard}.}

\begin{table}[htbp]
\centering
\scriptsize
\resizebox{\linewidth}{!}{%
\setlength{\tabcolsep}{2pt}
    \begin{tabular}{l|cccccc|c|cccccc|c}
    \toprule
    & \multicolumn{7}{c|}{\textbf{Zero-shot}}
    & \multicolumn{7}{c}{\textbf{Few-shot} (5-shot; 10-shot for HellaSwag)} \\
    \cmidrule(lr){2-8} \cmidrule(lr){9-15}
    \textbf{Model}
    & \textbf{\begin{sideways}Kyrgyz\textsubscript{MMLU}\end{sideways}}
    & \textbf{\begin{sideways}Kyrgyz\textsubscript{RC}\end{sideways}}
    & \textbf{\begin{sideways}Wino\allowbreak Grande\end{sideways}}
    & \textbf{\begin{sideways}BoolQ\end{sideways}}
    & \textbf{\begin{sideways}Hella\allowbreak Swag\end{sideways}}
    & \textbf{\begin{sideways}Truthful\allowbreak QA\end{sideways}}
    & \textbf{Avg}
    & \textbf{\begin{sideways}Kyrgyz\textsubscript{MMLU}\end{sideways}}
    & \textbf{\begin{sideways}Kyrgyz\textsubscript{RC}\end{sideways}}
    & \textbf{\begin{sideways}Wino\allowbreak Grande\end{sideways}}
    & \textbf{\begin{sideways}BoolQ\end{sideways}}
    & \textbf{\begin{sideways}Hella\allowbreak Swag\end{sideways}}
    & \textbf{\begin{sideways}Truthful\allowbreak QA\end{sideways}}
    & \textbf{Avg} \\
    \midrule
    Qwen2.5-0.5B-Instruct & 27.4 & 53.2 & \textbf{51.5} & 37.9 & 14.6 & 33.5 & 36.4 & \cellcolor{red!12}25.4 & \cellcolor{green!12}54.0 & \cellcolor{red!12}49.7 & \cellcolor{green!32}61.0 & \cellcolor{green!36}25.9 & \cellcolor{red!12}33.4 & \cellcolor{green!41}41.6 \\
    Qwen2.5-1.5B-Instruct & 27.9 & 60.5 & 50.1 & 38.6 & 22.9 & 32.5 & 38.8 & \cellcolor{green!12}28.7 & \cellcolor{green!32}67.5 & 50.1 & \cellcolor{green!28}58.0 & \cellcolor{green!16}26.5 & \cellcolor{green!12}32.9 & \cellcolor{green!40}43.9 \\
    Qwen2.5-3B-Instruct & 28.6 & 66.0 & 50.5 & \textbf{59.4} & 22.0 & 34.2 & 43.4 & \cellcolor{green!40}34.0 & \cellcolor{green!40}73.2 & \cellcolor{green!12}51.3 & \cellcolor{red!12}57.4 & \cellcolor{green!12}23.7 & \cellcolor{green!12}34.4 & \cellcolor{green!23}45.7 \\
    Qwen2.5-7B-Instruct & 31.5 & 70.0 & 48.7 & 56.3 & 10.0 & 34.1 & 41.8 & \cellcolor{green!40}38.5 & \cellcolor{green!20}74.8 & \cellcolor{green!12}50.4 & \cellcolor{green!20}64.6 & \cellcolor{green!30}17.8 & \cellcolor{green!20}36.2 & \cellcolor{green!41}47.1 \\
    \midrule
    Qwen3-0.6B & 26.0 & 61.8 & 49.8 & 38.0 & 11.1 & 29.9 & 36.1 & \cellcolor{green!12}26.8 & \cellcolor{red!12}59.5 & \cellcolor{green!12}50.1 & \cellcolor{green!32}60.1 & \cellcolor{green!36}26.4 & \cellcolor{green!12}30.0 & \cellcolor{green!46}42.2 \\
    Qwen3-1.7B & 27.9 & 61.8 & 48.9 & 40.4 & 24.6 & 29.6 & 38.9 & \cellcolor{green!20}30.8 & \cellcolor{green!40}71.2 & \cellcolor{red!12}48.6 & \cellcolor{green!32}62.0 & \cellcolor{green!12}25.2 & \cellcolor{green!12}30.3 & \cellcolor{green!44}44.7 \\
    Qwen3-4B & 30.3 & 68.2 & 49.0 & 38.3 & 24.5 & 32.9 & 40.5 & \cellcolor{green!40}38.5 & \cellcolor{green!40}77.2 & \cellcolor{red!12}48.1 & \cellcolor{green!60}74.0 & \cellcolor{green!12}24.7 & \cellcolor{red!12}32.5 & \cellcolor{green!62}49.2 \\
    Qwen3-8B & \textbf{32.1} & 71.8 & 51.0 & 39.2 & 24.6 & \textbf{34.7} & 42.2 & \cellcolor{green!40}\textbf{44.5} & \cellcolor{green!40}\textbf{81.8} & \cellcolor{red!12}50.6 & \cellcolor{green!60}\textbf{76.9} & \cellcolor{green!12}26.4 & \cellcolor{green!12}35.8 & \cellcolor{green!70}\textbf{52.7} \\
    \midrule
    Gemma-3-270m & 27.5 & 56.8 & 48.3 & 37.9 & 17.4 & \textbf{34.7} & 37.1 & \cellcolor{red!12}27.0 & \cellcolor{red!12}53.2 & \cellcolor{green!12}48.7 & \cellcolor{green!40}61.5 & \cellcolor{green!40}\textbf{27.6} & \cellcolor{green!12}36.6 & \cellcolor{green!41}42.4 \\
    Gemma-3-1b-it & 26.7 & 58.2 & 50.0 & 37.9 & 24.4 & 34.0 & 38.5 & \cellcolor{red!12}26.5 & \cellcolor{red!70}38.0 & \cellcolor{red!12}48.9 & \cellcolor{green!40}62.8 & \cellcolor{red!12}23.5 & \cellcolor{red!12}31.3 & 38.5 \\
    Gemma-3-4b-it & 30.3 & 70.2 & 50.6 & 58.3 & 24.6 & \textbf{34.7} & \textbf{44.8} & \cellcolor{red!12}29.5 & \cellcolor{red!70}25.0 & \cellcolor{red!12}49.6 & \cellcolor{green!12}62.1 & 24.6 & \cellcolor{green!70}\textbf{50.0} & \cellcolor{red!38}40.1 \\
    \midrule
    Llama-3.1-8B-Instruct & 31.0 & \textbf{75.2} & 50.6 & 50.3 & \textbf{26.6} & 33.7 & 44.6 & \cellcolor{green!40}38.1 & \cellcolor{green!12}80.5 & \cellcolor{green!12}\textbf{51.6} & \cellcolor{green!40}75.5 & \cellcolor{red!12}21.9 & \cellcolor{green!12}34.4 & \cellcolor{green!44}50.3 \\
    Llama-3.2-1B-Instruct & 26.3 & 58.2 & 49.4 & 38.3 & 0.2 & 30.1 & 33.7 & \cellcolor{red!12}26.1 & \cellcolor{red!12}45.8 & \cellcolor{green!12}49.7 & \cellcolor{green!40}62.0 & \cellcolor{green!40}25.8 & \cellcolor{green!12}30.3 & \cellcolor{green!47}40.0 \\
    Llama-3.2-3B-Instruct & 27.8 & 64.2 & 49.1 & 43.1 & 24.5 & 31.5 & 40.0 & \cellcolor{green!12}29.4 & \cellcolor{green!12}64.8 & \cellcolor{red!12}48.9 & \cellcolor{green!40}62.3 & \cellcolor{green!12}25.3 & \cellcolor{green!12}32.9 & \cellcolor{green!33}43.9 \\
    \bottomrule
    \end{tabular}%
}
\caption{Zero-shot and few-shot evaluation results on Kyrgyz benchmarks (accuracy \%). Cell colors in the few-shot half indicate gains/losses relative to the zero-shot result for the same model and metric.}
\label{tab:kg_combined_results}
\end{table}

\textbf{Open-source models.}
We evaluated $14$ open-source models from  Qwen~\cite{bai2023qwentechnicalreport}, Gemma~\cite{gemmateam2024gemmaopenmodelsbased}, and LLaMA~\cite{grattafiori2024llama3herdmodels} families in zero-shot and few-shot settings.
Inference was performed on rented NVIDIA RTX6000 Ada and NVIDIA L40S GPUs.

Few-shot evaluation used 5 examples for most tasks.
For \emph{HellaSwag}, we use 10-shot prompting, following common practice in widely adopted benchmarking frameworks and leaderboards.

Responses were parsed using a standard \textit{Lighteval} regex to extract answers, and accuracy was reported as the percentage of correct responses.
For open models, we additionally report parallel English baselines in Table~\ref{tab:en_combined_results} as a cross-lingual reference.
A detailed analysis is provided in Sections~\ref{sec:evaluation-results}--\ref{sec:discussion}.

\textbf{Proprietary models.}
To assess state-of-the-art closed-source model performance on Kyrgyz in a cost-effective yet representative manner, we conducted an evaluation using a condensed benchmark, \emph{KyrgyzLLM Tiny Bench}.
It consists of $100$ randomly selected questions per subject from the original suite, sampled with a fixed random seed for reproducibility.
Evaluated models include GPT-5.1 and GPT-5 Mini (OpenAI); Claude 4.5 Sonnet and Haiku (Anthropic); Gemini 2.5 Flash (Google); Grok 4.1 Fast (xAI); Mistral Medium 3.1 and Large (Mistral); DeepSeek V3.2 Exp; Qwen3-Max; Kimi K2 Turbo; and GigaChat-2 Max.
Results for Gemini 2.5 Flash (marked *) were affected by safety-related refusals; the reported scores should therefore be treated with caution. The in-context supervision strategy matched that used for open-source models.

    \setlength{\tabcolsep}{2pt}
    \begin{sidewaystable}[htbp]
    \centering
    \footnotesize
    \resizebox{\dimexpr\textheight-2\baselineskip}{!}{%
    \begin{tabular}{l|ccccccccc|c|cccc|c|cccc|c}
    \toprule
    & \multicolumn{10}{c|}{\textbf{MMLU}} & \multicolumn{5}{c|}{\textbf{RC}} & \multicolumn{4}{c|}{{\textbf{Translated}}} & \textbf{Total} \\
    \textbf{Model} & \begin{sideways}Bio\end{sideways} & \begin{sideways}Chem\end{sideways} & \begin{sideways}Geog\end{sideways} & \begin{sideways}Hist\end{sideways} &\begin{sideways}Lang\end{sideways} & \begin{sideways}Lit\end{sideways} & \begin{sideways}Math\end{sideways} & \begin{sideways}Med\end{sideways} & \begin{sideways}Phys\end{sideways} & {\textbf{Avg}} & \begin{sideways}Lit\end{sideways} & \begin{sideways}Math\end{sideways} & \begin{sideways}News\end{sideways} & \begin{sideways}Wiki\end{sideways} & \textbf{Avg} & \textbf{BoolQ} & \textbf{Hella} & \textbf{TQA} & \textbf{Wino} & {\textbf{Avg}} \\
    \midrule
    \multicolumn{21}{c}{\textbf{Zero-shot evaluation}} \\
    \midrule
    GigaChat-2-Max
    & 64.0 & 60.0 & 60.0 & 70.0 & 55.0 & 43.0 & 49.0 & 64.0 & 51.0 & 57.33 & 89.0 & 66.0 & 87.0 & 99.0 & 85.25 & 83.0 & 48.0 & 44.0 & 48.0 & 63.76 \\

    Claude 4.5 Haiku
    & 53.0 & 61.0 & 52.0 & 69.0 & 71.0 & 36.0 & 54.0 & 60.0 & 76.0 & 59.11 & 92.0 & 88.0 & 86.0 & 98.0 & 91.00 & 87.0 & 0.0 & 61.0 & 48.0 & 64.06 \\

    Claude 4.5 Sonnet
    & 61.0 & 69.0 & 57.0 & 81.0 & 82.0 & 49.0 & 53.0 & 72.0 & 77.0 & \textbf{66.78} & 93.0 & 98.0 & 89.0 & 99.0 & \textbf{94.75} & 95.0 & 77.0 & 69.0 & 51.0 & \textbf{74.82} \\

    DeepSeek V3.2 Exp
    & 60.0 & 65.0 & 63.0 & 83.0 & 71.0 & 43.0 & 53.0 & 61.0 & 77.0 & 64.00 & 94.0 & 84.0 & 84.0 & 98.0 & 90.00 & 84.0 & 56.0 & 66.0 & 56.0 & 70.47 \\

    \rowcolor{gray!10} Gemini 2.5 Flash~*
    & 43.0 & 42.0 & 45.0 & 57.0 & 85.0 & 31.0 & 50.0 & 43.0 & 54.0 & 50.00 & 96.0 & 81.0 & 82.0 & 95.0 & 88.50 & 79.0 & 34.0 & 4.0 & 49.0 & 57.06 \\

    GPT-5 Mini
    & 50.0 & 39.0 & 61.0 & 70.0 & 55.0 & 38.0 & 43.0 & 53.0 & 42.0 & 50.11 & 88.0 & 66.0 & 84.0 & 95.0 & 83.25 & 86.0 & 59.0 & 62.0 & 49.0 & 61.35 \\

    GPT-5.1
    & 62.0 & 54.0 & 67.0 & 82.0 & 77.0 & 46.0 & 37.0 & 63.0 & 48.0 & 59.56 & 94.0 & 85.0 & 89.0 & 98.0 & 91.50 & 88.0 & 79.0 & 68.0 & 55.0 & 70.12 \\

    Grok 4.1 Fast
    & 43.0 & 50.0 & 64.0 & 67.0 & 44.0 & 41.0 & 32.0 & 56.0 & 40.0 & 48.56 & 92.0 & 64.0 & 87.0 & 100 & 85.75 & 86.0 & 59.0 & 48.0 & 56.0 & 60.53 \\

    {Kimi K2 Turbo}
    & 58.0 & 54.0 & 63.0 & 71.0 & 70.0 & 40.0 & 42.0 & 58.0 & 55.0 & 56.78 & 94.0 & 70.0 & 89.0 & 99.0 & 88.00 & 86.0 & 59.0 & 50.0 & 58.0 & 65.70 \\

    Mistral Large
    & 60.0 & 65.0 & 63.0 & 78.0 & 63.0 & 38.0 & 50.0 & 62.0 & 74.0 & 61.44 & 94.0 & 81.0 & 89.0 & 98.0 & 90.50 & 86.0 & 64.0 & 56.0 & 52.0 & 68.94 \\

    Mistral Medium 3.1
    & 46.0 & 54.0 & 64.0 & 67.0 & 47.0 & 45.0 & 41.0 & 63.0 & 55.0 & 53.56 & 94.0 & 68.0 & 88.0 & 98.0 & 87.00 & 76.0 & 56.0 & 50.0 & 50.0 & 62.47 \\

    Qwen3-Max
    & 57.0 & 63.0 & 61.0 & 72.0 & 72.0 & 46.0 & 49.0 & 59.0 & 57.0 & 59.56 & 92.0 & 85.0 & 88.0 & 99.0 & 91.00 & 90.0 & 66.0 & 65.0 & 53.0 & 69.06 \\

    \midrule
    \multicolumn{21}{c}{\textbf{Few-shot evaluation}} \\
    \midrule
    GigaChat-2-Max
    & \cellcolor{green!12}65.0 & \cellcolor{red!12}57.0 & \cellcolor{green!12}65.0 & \cellcolor{green!12}77.0 & \cellcolor{green!16}72.0 & \cellcolor{green!12}44.0 & \cellcolor{red!12}46.0 & 64.0 & \cellcolor{red!12}47.0 & \cellcolor{green!12}59.67 & 89.0 & \cellcolor{red!12}65.0 & \cellcolor{green!12}89.0 & \cellcolor{red!12}94.0 & \cellcolor{red!12}84.25 & \cellcolor{red!12}69.0 & \cellcolor{red!12}45.0 & \cellcolor{red!12}36.0 & \cellcolor{green!12}50.0 & \cellcolor{green!12}63.82 \\

    Claude 4.5 Haiku
    & \cellcolor{green!12}65.0 & \cellcolor{green!12}62.0 & \cellcolor{green!12}63.0 & \cellcolor{green!12}75.0 & \cellcolor{green!12}78.0 & \cellcolor{green!12}37.0 & \cellcolor{red!12}53.0 & \cellcolor{red!12}55.0 & \cellcolor{red!12}74.0 & \cellcolor{green!12}62.44 & \cellcolor{red!12}67.0 & \cellcolor{red!12}86.0 & \cellcolor{red!12}72.0 & \cellcolor{red!28}57.0 & \cellcolor{red!12}70.50 & \cellcolor{red!70}14.0 & \cellcolor{green!12}62.0 & \cellcolor{green!12}74.0 & \cellcolor{green!12}50.0 & \cellcolor{red!13}62.59 \\

    Claude 4.5 Sonnet
    & \cellcolor{green!12}63.0 & \cellcolor{red!12}68.0 & \cellcolor{green!12}68.0 & \cellcolor{green!12}84.0 & \cellcolor{green!12}84.0 & \cellcolor{green!12}51.0 & 53.0 & \cellcolor{red!12}70.0 & \cellcolor{red!12}76.0 & \cellcolor{green!12}\textbf{68.56} & \cellcolor{red!12}92.0 & \cellcolor{red!12}97.0 & \cellcolor{green!12}91.0 & 99.0 & \textbf{94.75} & \cellcolor{red!12}79.0 & \cellcolor{green!12}84.0 & \cellcolor{green!24}96.0 & \cellcolor{green!12}55.0 & \cellcolor{green!12}\textbf{77.06} \\

    DeepSeek V3.2 Exp
    & \cellcolor{green!12}63.0 & \cellcolor{red!12}62.0 & \cellcolor{green!12}64.0 & \cellcolor{red!12}82.0 & \cellcolor{red!12}61.0 & \cellcolor{green!12}47.0 & \cellcolor{red!12}49.0 & \cellcolor{red!12}54.0 & \cellcolor{red!21}55.0 & \cellcolor{red!12}59.67 & 94.0 & \cellcolor{green!12}86.0 & \cellcolor{green!12}92.0 & \cellcolor{green!12}100.0 & \cellcolor{green!12}93.00 & \cellcolor{red!12}70.0 & \cellcolor{red!20}35.0 & \cellcolor{red!12}59.0 & \cellcolor{red!12}48.0 & \cellcolor{red!12}61.01 \\

    \rowcolor{gray!10} Gemini 2.5 Flash~*
    & \cellcolor{green!13}57.0 & \cellcolor{green!18}61.0 & \cellcolor{green!27}74.0 & \cellcolor{green!21}79.0 & \cellcolor{green!12}89.0 & \cellcolor{green!21}52.0 & \cellcolor{red!12}48.0 & \cellcolor{green!18}63.0 & \cellcolor{green!15}69.0 & \cellcolor{green!19}65.78 & \cellcolor{green!12}97.0 & \cellcolor{green!12}88.0 & \cellcolor{red!12}78.0 & 95.0 & \cellcolor{green!12}89.50 & \cellcolor{red!28}47.0 & \cellcolor{green!12}38.0 & \cellcolor{green!16}21.0 & \cellcolor{red!12}47.0 & \cellcolor{green!12}59.88 \\

    GPT-5 Mini
    & \cellcolor{green!12}58.0 & \cellcolor{green!25}66.0 & \cellcolor{green!12}73.0 & \cellcolor{green!12}82.0 & \cellcolor{green!24}81.0 & \cellcolor{red!12}36.0 & \cellcolor{green!12}52.0 & \cellcolor{green!12}58.0 & \cellcolor{green!26}77.0 & \cellcolor{green!13}64.78 & \cellcolor{green!12}91.0 & \cellcolor{green!25}93.0 & \cellcolor{green!12}89.0 & \cellcolor{green!12}98.0 & \cellcolor{green!12}92.75 & \cellcolor{red!42}41.0 & \cellcolor{red!12}45.0 & \cellcolor{red!30}26.0 & \cellcolor{red!12}48.0 & \cellcolor{red!12}59.74 \\

    GPT-5.1
    & \cellcolor{green!12}65.0 & \cellcolor{green!12}62.0 & \cellcolor{green!12}68.0 & \cellcolor{green!12}83.0 & \cellcolor{green!12}85.0 & \cellcolor{red!12}44.0 & \cellcolor{green!12}41.0 & \cellcolor{green!12}64.0 & \cellcolor{green!12}51.0 & \cellcolor{green!12}62.56 & \cellcolor{red!12}93.0 & \cellcolor{red!12}84.0 & \cellcolor{green!12}90.0 & 98.0 & \cellcolor{red!12}91.25 & \cellcolor{red!70}13.0 & 79.0 & \cellcolor{red!25}41.0 & \cellcolor{green!12}58.0 & \cellcolor{red!12}68.84 \\

    Grok 4.1 Fast
    & \cellcolor{green!12}51.0 & \cellcolor{green!12}52.0 & \cellcolor{green!12}67.0 & \cellcolor{green!12}71.0 & \cellcolor{green!12}56.0 & 41.0 & \cellcolor{green!12}36.0 & \cellcolor{green!12}58.0 & \cellcolor{green!12}50.0 & \cellcolor{green!12}53.56 & \cellcolor{green!12}93.0 & \cellcolor{red!12}58.0 & \cellcolor{green!12}90.0 & 100.0 & \cellcolor{red!12}85.25 & \cellcolor{red!66}16.0 & \cellcolor{red!12}50.0 & \cellcolor{green!12}67.0 & \cellcolor{red!12}53.0 & \cellcolor{red!12}59.35 \\

    {Kimi K2 Turbo}
    & 58.0 & \cellcolor{green!12}57.0 & 63.0 & \cellcolor{green!12}72.0 & \cellcolor{green!12}72.0 & \cellcolor{green!12}47.0 & 42.0 & \cellcolor{green!12}60.0 & \cellcolor{red!12}48.0 & \cellcolor{green!12}57.67 & \cellcolor{green!12}96.0 & \cellcolor{red!12}68.0 & \cellcolor{green!12}92.0 & \cellcolor{green!12}100.0 & \cellcolor{green!12}89.00 & \cellcolor{red!12}79.0 & \cellcolor{green!18}70.0 & \cellcolor{green!24}79.0 & \cellcolor{red!12}51.0 & \cellcolor{green!12}67.88 \\

    Mistral Large
    & \cellcolor{green!12}61.0 & \cellcolor{red!12}56.0 & \cellcolor{green!12}69.0 & \cellcolor{green!12}84.0 & \cellcolor{green!12}75.0 & \cellcolor{green!12}41.0 & \cellcolor{red!12}41.0 & \cellcolor{green!12}68.0 & \cellcolor{red!18}56.0 & \cellcolor{red!12}61.22 & \cellcolor{red!12}93.0 & \cellcolor{red!12}78.0 & \cellcolor{green!12}91.0 & \cellcolor{green!12}99.0 & \cellcolor{red!12}90.25 & \cellcolor{red!22}63.0 & \cellcolor{red!18}35.0 & \cellcolor{red!12}53.0 & \cellcolor{green!12}57.0 & \cellcolor{red!12}61.65 \\

    Mistral Medium 3.1
    & \cellcolor{green!12}57.0 & \cellcolor{red!12}52.0 & \cellcolor{green!12}68.0 & \cellcolor{green!12}76.0 & \cellcolor{green!12}58.0 & \cellcolor{green!12}48.0 & \cellcolor{green!12}42.0 & \cellcolor{green!12}64.0 & \cellcolor{green!12}67.0 & \cellcolor{green!12}59.11 & \cellcolor{green!12}95.0 & \cellcolor{green!12}79.0 & \cellcolor{green!12}90.0 & \cellcolor{green!12}100.0 & \cellcolor{green!12}91.00 & \cellcolor{red!55}17.0 & \cellcolor{red!12}46.0 & \cellcolor{red!12}49.0 & \cellcolor{green!12}53.0 & \cellcolor{red!12}62.29 \\

    Qwen3-Max
    & \cellcolor{green!12}65.0 & \cellcolor{green!12}64.0 & \cellcolor{green!12}64.0 & \cellcolor{green!12}79.0 & \cellcolor{green!12}76.0 & \cellcolor{red!12}42.0 & \cellcolor{red!12}46.0 & \cellcolor{green!12}61.0 & 57.0 & \cellcolor{green!12}61.56 & \cellcolor{red!12}91.0 & \cellcolor{green!12}87.0 & 88.0 & \cellcolor{green!12}100.0 & \cellcolor{green!12}91.50 & \cellcolor{red!12}73.0 & \cellcolor{red!12}63.0 & \cellcolor{green!26}98.0 & \cellcolor{red!12}50.0 & \cellcolor{green!12}70.82 \\

    \bottomrule
    \end{tabular}
    }
    \caption{{Zero-shot} and few-shot accuracy (\%) on {\emph{KyrgyzLLM Tiny Bench}} (100 sample subset). * Gemini 2.5 Flash scores were impacted by safety filter refusals.}
    \label{tab:proprietary_combined}
    \end{sidewaystable}

\section{Evaluation Results}\label{sec:evaluation-results}

We report accuracy (\%) and macro-averages (arithmetic mean).
Because tasks probe different LLM capabilities, averages over all tasks should be interpreted as coarse summaries rather than definitive diagnostic scores.

\textbf{Open-source models.}
Table~\ref{tab:kg_combined_results} shows consistent within-family scaling, with larger instruction-tuned models performing best.
In few-shot Kyrgyz evaluation, {Qwen3-8B} achieves the highest average accuracy (52.7\%), followed by \emph{Llama-3.1-8B-Instruct} (50.3\%).
The largest few-shot gains are observed on \emph{BoolQ} ({Qwen3-8B}: 39.2$\rightarrow$76.9).
English baselines {(Table~\ref{tab:en_combined_results})} show {Qwen2.5-7B-Instruct} leading both zero-shot (67.9\%) and few-shot (70.1\%).
{Per-subject breakdowns on \textit{KyrgyzMMLU} and per-genre breakdowns on \textit{KyrgyzRC} are reported in Tables~\ref{tab:ky_mmlu_zs_and_delta} and~\ref{tab:ky_rc_zs_and_delta}, respectively.}

\textbf{Proprietary models.} Table~\ref{tab:proprietary_combined} shows \emph{Claude~4.5~Sonnet} achieving the highest average accuracy in both zero-shot (74.82\%) and few-shot (77.06\%).
Performance is near ceiling on reading comprehension, particularly \textit{RC\textsubscript{Wiki}} (98--100\% for several models), whereas science-heavy \textit{KyrgyzMMLU} subjects exhibit substantially higher variance, suggesting that out-of-context factual and numerical reasoning in Kyrgyz remains more challenging than extracting answers from a given passage.
Few-shot prompting often improves results, but gains are model- and task-dependent.

\begin{sidewaystable}[htbp]
\centering
\footnotesize
\resizebox{\dimexpr\textheight-2\baselineskip\relax}{!}{%
    \begin{tabular}{l*{9}{r r}|l*{1}{r r}}
    \toprule
    \textbf{Model} & \multicolumn{2}{c}{\textbf{Medicine}} & \multicolumn{2}{c}{\textbf{History}} & \multicolumn{2}{c}{\textbf{Literature}} & \multicolumn{2}{c}{\textbf{Lang}} & \multicolumn{2}{c}{\textbf{Biology}} & \multicolumn{2}{c}{\textbf{Chemistry}} & \multicolumn{2}{c}{\textbf{Math}} & \multicolumn{2}{c}{\textbf{Physics}} & \multicolumn{2}{c}{\textbf{Geography}} & \multicolumn{2}{c}{\textbf{Average}} \\
    \cmidrule(lr){2-3} \cmidrule(lr){4-5} \cmidrule(lr){6-7} \cmidrule(lr){8-9} \cmidrule(lr){10-11} \cmidrule(lr){12-13} \cmidrule(lr){14-15} \cmidrule(lr){16-17} \cmidrule(lr){18-19} \cmidrule(lr){20-21}
     & \textbf{ZS} & \textbf{$\Delta$} & \textbf{ZS} & \textbf{$\Delta$} & \textbf{ZS} & \textbf{$\Delta$} & \textbf{ZS} & \textbf{$\Delta$} & \textbf{ZS} & \textbf{$\Delta$} & \textbf{ZS} & \textbf{$\Delta$} & \textbf{ZS} & \textbf{$\Delta$} & \textbf{ZS} & \textbf{$\Delta$} & \textbf{ZS} & \textbf{$\Delta$} & \textbf{ZS} & \textbf{$\Delta$} \\
    \midrule
    Qwen2.5-0.5B-Instruct & 31.5 & \cellcolor{red!10}-2.3 & 33.9 & \cellcolor{red!10}-2.3 & 26.3 & \cellcolor{red!10}-2.4 & 27.2 & \cellcolor{red!25}-8.3 & 30.8 & \cellcolor{red!10}-2.8 & 20.7 & \cellcolor{green!10}+2.9 & 22.8 & \cellcolor{green!10}+1.4 & 26.1 & \cellcolor{red!10}-0.4 & 26.9 & \cellcolor{red!10}-3.8 & 27.4 & \cellcolor{red!10}-2.0 \\
    Qwen2.5-1.5B-Instruct & 29.6 & \cellcolor{green!11}+3.7 & 35.9 & \cellcolor{red!12}-4.1 & 27.7 & -0.1 & 29.4 & \cellcolor{red!20}-6.6 & 31.8 & \cellcolor{red!10}-1.4 & 20.5 & \cellcolor{green!17}+5.6 & 25.1 & \cellcolor{green!14}+4.8 & 26.6 & \cellcolor{green!10}+2.1 & 24.8 & \cellcolor{green!12}+3.2 & 27.9 & \cellcolor{green!10}+0.8 \\
    Qwen2.5-3B-Instruct & 32.9 & \cellcolor{green!18}+6.0 & 36.8 & \cellcolor{green!10}+3.4 & 27.7 & \cellcolor{green!10}+3.2 & 26.7 & \cellcolor{green!10}+0.8 & 32.1 & \cellcolor{green!13}+4.2 & 23.0 & \cellcolor{green!22}+7.4 & 26.9 & \cellcolor{green!16}+5.3 & 26.5 & \cellcolor{green!14}+4.6 & 25.0 & \cellcolor{green!32}+10.5 & 28.6 & \cellcolor{green!15}+5.1 \\
    Qwen2.5-7B-Instruct & \textbf{35.2} & \cellcolor{green!14}+4.6 & 40.7 & \cellcolor{green!12}+4.1 & 29.1 & \cellcolor{green!10}+1.7 & 25.3 & \cellcolor{green!32}+10.8 & \textbf{34.1} & \cellcolor{green!17}+5.6 & 26.0 & \cellcolor{green!34}+11.4 & 31.8 & \cellcolor{green!19}+6.4 & \textbf{30.3} & \cellcolor{green!28}+9.2 & 30.9 & \cellcolor{green!28}+9.4 & 31.5 & \cellcolor{green!21}+7.0 \\
    Qwen3-0.6B & 28.2 & \cellcolor{green!10}+1.9 & 33.4 & +0.0 & 27.1 & \cellcolor{green!10}+1.6 & 26.1 & \cellcolor{red!10}-3.3 & 30.4 & \cellcolor{red!10}-2.8 & 19.5 & \cellcolor{green!16}+5.3 & 21.4 & \cellcolor{green!15}+5.0 & 24.9 & \cellcolor{red!10}-1.2 & 23.1 & \cellcolor{green!10}+0.8 & 26.0 & \cellcolor{green!10}+0.8 \\
    Qwen3-1.7B & 29.6 & \cellcolor{green!10}+1.0 & 36.4 & \cellcolor{red!10}-1.2 & 28.3 & \cellcolor{green!10}+3.3 & 25.3 & \cellcolor{red!20}-6.7 & 30.7 & \cellcolor{green!17}+5.6 & 22.1 & \cellcolor{green!30}+10.1 & 26.9 & \cellcolor{green!12}+4.0 & 25.7 & \cellcolor{green!10}+2.3 & 26.1 & \cellcolor{green!22}+7.5 & 27.9 & \cellcolor{green!10}+2.9 \\
    Qwen3-4B & 28.7 & \cellcolor{green!28}+9.3 & 36.1 & \cellcolor{green!14}+4.8 & 28.0 & \cellcolor{green!11}+3.7 & 27.5 & \cellcolor{green!10}+0.3 & 32.8 & \cellcolor{green!25}+8.3 & \textbf{28.0} & \cellcolor{green!51}+17.2 & 32.1 & \cellcolor{green!14}+4.8 & 29.9 & \cellcolor{green!40}+13.4 & 29.5 & \cellcolor{green!38}+12.5 & 30.3 & \cellcolor{green!25}+8.2 \\
    Qwen3-8B & 32.4 & \cellcolor{green!31}+10.2 & 42.5 & \cellcolor{green!32}+10.7 & 29.8 & \cellcolor{green!10}+3.2 & 28.9 & \cellcolor{green!18}+6.1 & 33.6 & \cellcolor{green!46}+15.3 & 27.4 & \cellcolor{green!60}+21.2 & \textbf{34.0} & \cellcolor{green!20}+6.8 & 28.8 & \cellcolor{green!54}+18.0 & 31.7 & \cellcolor{green!59}+19.7 & \textbf{32.1} & \cellcolor{green!37}+12.4 \\
    \midrule
    Gemma-3-270m & 32.4 & \cellcolor{red!22}-7.4 & 31.6 & \cellcolor{red!17}-5.7 & 27.9 & \cellcolor{green!10}+1.2 & \textbf{34.7} & \cellcolor{green!20}+6.7 & 30.4 & \cellcolor{red!10}-1.0 & 19.3 & \cellcolor{green!10}+2.4 & 21.8 & \cellcolor{red!10}-0.2 & 24.8 & \cellcolor{red!10}-2.6 & 24.4 & \cellcolor{green!10}+2.0 & 27.5 & \cellcolor{red!10}-0.5 \\
    Gemma-3-1b-it & 27.8 & \cellcolor{green!10}+0.9 & 34.3 & \cellcolor{red!13}-4.5 & 27.0 & \cellcolor{red!10}-1.1 & 30.8 & \cellcolor{green!10}+0.3 & 29.4 & \cellcolor{green!10}+0.7 & 21.7 & \cellcolor{green!10}+0.4 & 22.0 & \cellcolor{red!10}-2.1 & 24.1 & \cellcolor{green!10}+0.8 & 23.6 & \cellcolor{green!10}+2.3 & 26.7 & \cellcolor{red!10}-0.2 \\
    Gemma-3-4b-it & 33.3 & \cellcolor{green!17}+5.6 & 39.8 & \cellcolor{green!11}+3.6 & \textbf{29.9} & \cellcolor{red!10}-1.2 & 29.4 & \cellcolor{green!34}+11.4 & 32.6 & \cellcolor{red!16}-5.4 & 23.3 & \cellcolor{red!10}-2.8 & 28.1 & \cellcolor{red!27}-9.1 & 28.3 & \cellcolor{red!16}-5.2 & 28.3 & \cellcolor{red!13}-4.4 & 30.3 & \cellcolor{red!10}-0.8 \\
    \midrule
    Llama-3.1-8B-Instruct & \textbf{35.2} & \cellcolor{green!10}+1.8 & \textbf{43.9} & \cellcolor{green!13}+4.5 & 29.4 & \cellcolor{green!10}+0.9 & 25.0 & \cellcolor{green!39}+13.1 & 32.8 & \cellcolor{green!31}+10.5 & 24.9 & \cellcolor{green!32}+10.5 & 27.7 & \cellcolor{green!14}+4.7 & 27.2 & \cellcolor{green!18}+5.9 & \textbf{32.5} & \cellcolor{green!38}+12.7 & 31.0 & \cellcolor{green!21}+7.1 \\
    Llama-3.2-1B-Instruct & 26.9 & \cellcolor{green!10}+0.9 & 29.8 & \cellcolor{red!12}-4.1 & 27.8 & \cellcolor{green!10}+1.0 & 28.9 & +0.0 & 30.8 & \cellcolor{red!13}-4.3 & 21.8 & \cellcolor{green!10}+1.4 & 22.0 & \cellcolor{green!10}+2.1 & 25.2 & \cellcolor{red!10}-1.6 & 23.4 & \cellcolor{green!10}+3.2 & 26.3 & \cellcolor{red!10}-0.2 \\
    Llama-3.2-3B-Instruct & 27.3 & \cellcolor{green!10}+2.3 & 34.8 & \cellcolor{red!10}-2.1 & 27.8 & \cellcolor{red!10}-1.2 & 25.3 & \cellcolor{green!10}+2.8 & 29.6 & \cellcolor{green!10}+2.9 & 23.7 & \cellcolor{green!10}+2.8 & 26.7 & \cellcolor{green!10}+1.6 & 25.2 & \cellcolor{green!10}+1.3 & 29.5 & \cellcolor{green!14}+4.6 & 27.8 & \cellcolor{green!10}+1.6 \\
    \bottomrule
    \end{tabular}%
}
\caption{Zero-shot accuracy on \textit{KyrgyzMMLU} (ZS, \%); few-shot change ($\Delta$ = few-shot $-$ zero-shot, \% points).}
\label{tab:ky_mmlu_zs_and_delta}
\end{sidewaystable}

\begin{table}[htbp]
\centering
\setlength{\tabcolsep}{2pt}
\begin{tabular}{l*{4}{r r}|l*{1}{r r}}
\toprule
\textbf{Model}
& \multicolumn{2}{c}{\textbf{Lit}}
& \multicolumn{2}{c}{\textbf{Math}}
& \multicolumn{2}{c}{\textbf{News}}
& \multicolumn{2}{c}{\textbf{Wiki}}
& \multicolumn{2}{c}{\textbf{Avg}} \\
\cmidrule(lr){2-3}\cmidrule(lr){4-5}\cmidrule(lr){6-7}\cmidrule(lr){8-9}\cmidrule(lr){10-11}
& \textbf{ZS} & \textbf{$\Delta$}
& \textbf{ZS} & \textbf{$\Delta$}
& \textbf{ZS} & \textbf{$\Delta$}
& \textbf{ZS} & \textbf{$\Delta$}
& \textbf{ZS} & \textbf{$\Delta$} \\
\midrule

Qwen2.5-0.5B-Instruct & 67.0 & \cellcolor{green!34}+12.0 & 32.0 & \cellcolor{red!16}-3.0 & 44.0 & 0.0 & 70.0 & \cellcolor{red!22}-6.0 & 53.3 & \cellcolor{green!11}+0.7 \\
Qwen2.5-1.5B-Instruct & 77.0 & \cellcolor{green!26}+8.0 & 45.0 & \cellcolor{red!24}-7.0 & 48.0 & \cellcolor{green!54}+22.0 & 72.0 & \cellcolor{green!20}+5.0 & 60.5 & \cellcolor{green!24}+7.0 \\
Qwen2.5-3B-Instruct & 79.0 & \cellcolor{green!20}+5.0 & 50.0 & \cellcolor{red!22}-6.0 & 60.0 & \cellcolor{green!34}+12.0 & 75.0 & \cellcolor{green!46}+18.0 & 66.0 & \cellcolor{green!25}+7.3 \\
Qwen2.5-7B-Instruct & 81.0 & \cellcolor{green!20}+5.0 & 55.0 & \cellcolor{red!38}-14.0 & 61.0 & \cellcolor{green!40}+15.0 & 83.0 & \cellcolor{green!36}+13.0 & 70.0 & \cellcolor{green!20}+4.8 \\
Qwen3-0.6B & 74.0 & \cellcolor{green!14}+2.0 & 50.0 & \cellcolor{red!38}-14.0 & 52.0 & \cellcolor{green!20}+5.0 & 71.0 & \cellcolor{red!14}-2.0 & 61.8 & \cellcolor{red!15}-2.3 \\
Qwen3-1.7B & 66.0 & \cellcolor{green!36}+13.0 & 53.0 & \cellcolor{red!18}-4.0 & 61.0 & \cellcolor{green!32}+11.0 & 67.0 & \cellcolor{green!46}+18.0 & 61.8 & \cellcolor{green!29}+9.5 \\
Qwen3-4B & 80.0 & 0.0 & 54.0 & \cellcolor{green!14}+2.0 & 63.0 & \cellcolor{green!42}+16.0 & 76.0 & \cellcolor{green!46}+18.0 & 68.3 & \cellcolor{green!28}+9.0 \\
Qwen3-8B & 80.0 & \cellcolor{green!26}+8.0 & 66.0 & \cellcolor{red!16}-3.0 & 66.0 & \cellcolor{green!42}+16.0 & 75.0 & \cellcolor{green!48}+19.0 & 71.8 & \cellcolor{green!30}+10.0 \\
\midrule

Gemma-3-270m & 75.0 & \cellcolor{red!30}-10.0 & 28.0 & \cellcolor{red!36}-13.0 & 49.0 & \cellcolor{red!14}-2.0 & 75.0 & \cellcolor{green!32}+11.0 & 56.8 & \cellcolor{red!17}-3.5 \\
Gemma-3-1b-it & 79.0 & \cellcolor{red!70}-51.0 & 43.0 & \cellcolor{red!48}-19.0 & 45.0 & \cellcolor{red!54}-22.0 & 66.0 & \cellcolor{green!32}+11.0 & 58.3 & \cellcolor{red!51}-20.3 \\
Gemma-3-4b-it & 82.0 & \cellcolor{red!70}-82.0 & 50.0 & \cellcolor{red!70}-50.0 & 71.0 & \cellcolor{red!70}-71.0 & 78.0 & \cellcolor{green!54}+22.0 & 70.3 & \cellcolor{red!70}-45.3 \\
\midrule

Llama-3.1-8B-Instruct & 82.0 & \cellcolor{green!16}+3.0 & 65.0 & \cellcolor{red!24}-7.0 & 75.0 & \cellcolor{green!30}+10.0 & 79.0 & \cellcolor{green!40}+15.0 & 75.3 & \cellcolor{green!20}+5.2 \\
Llama-3.2-1B-Instruct & 71.0 & \cellcolor{red!60}-25.0 & 39.0 & \cellcolor{red!40}-15.0 & 53.0 & \cellcolor{red!44}-17.0 & 70.0 & \cellcolor{green!24}+7.0 & 58.3 & \cellcolor{red!35}-12.5 \\
Llama-3.2-3B-Instruct & 77.0 & \cellcolor{red!20}-5.0 & 45.0 & 0.0 & 62.0 & \cellcolor{red!28}-9.0 & 73.0 & \cellcolor{green!42}+16.0 & 64.3 & \cellcolor{green!11}+0.5 \\
\bottomrule
\end{tabular}%
\caption{Zero-shot accuracy on \textit{KyrgyzRC} (ZS, \%) and few-shot change ($\Delta$ = few-shot $-$ zero-shot, \% points).}
\label{tab:ky_rc_zs_and_delta}
\end{table}

\textbf{Scaling and few-shot effects.} Few-shot effects differ between open-source and proprietary models.
For open-source models, in-context demonstrations yield clear gains on \emph{KyrgyzRC} and \emph{BoolQ}, more moderate improvements on \emph{KyrgyzMMLU}, and limited gains on \emph{HellaSwag} (Table~\ref{tab:kg_combined_results}).
In contrast, proprietary models exhibit less consistent few-shot behavior: \emph{BoolQ} accuracy often stagnates or decreases relative to zero-shot performance, while gains on \emph{KyrgyzRC} are typically small due to already high zero-shot accuracy (Table~\ref{tab:proprietary_combined}).

\textbf{Cross-lingual consistency}. Model rankings in English and Kyrgyz are broadly preserved on \textit{WinoGrande/BoolQ}, and (to a lesser extent) on \textit{MMLU{/KyrgyzMMLU}}, indicating partial transferability of core reasoning and comprehension capabilities across languages.
In contrast, \textit{HellaSwag} exhibits the largest performance gap between English and Kyrgyz.
This pattern is consistent with the plausibility-shift hypothesis, which posits that translation-induced changes in discourse flow and event continuity disproportionately affect event-completion tasks.

\section{Discussion}\label{sec:discussion}

\textbf{Scaling trends.}
Across both Kyrgyz and English evaluations, performance consistently improves as model capacity increases within the same architectural family.
Higher-capacity and more extensively instruction-tuned variants outperform smaller counterparts across most subject areas.
These trends align with established scaling behaviors in multilingual benchmarks, indicating that model capacity and training quality remain key drivers even for less-resourced languages such as Kyrgyz.

\textbf{Effect of in-context learning (ICL).}
Few-shot prompting often improves results on \emph{KyrgyzRC} and \emph{BoolQ} for certain model families, where contextual examples aid comprehension and answer selection.
Observed gains of approximately $+5$–$+10$ points suggest that these datasets are suitable for assessing in-context adaptation.
In contrast, \emph{KyrgyzMMLU} shows more modest benefits.
Chain-of-thought prompting may further improve results but remains outside the scope of this study.

However, our results indicate that the effects of few-shot prompting are not uniform across tasks or models.
Few-shot benefits are strongly model- and task-dependent.
Using the \textit{Lighteval} framework with Kyrgyz-translated prompts, few-shot prompting occasionally reduces accuracy, suggesting that ICL effects in translated-prompt scenarios are more variable than often expected.
In particular, open-source models show substantial improvements on \emph{KyrgyzRC} and \emph{BoolQ} under in-context learning, whereas proprietary models—already strong in zero-shot settings—exhibit limited or even negative gains, especially on \emph{BoolQ}. For reading comprehension, performance saturation limits observable few-shot improvements among closed models.

The pronounced degradation on \textit{HellaSwag} provides empirical support for the plausibility-shift hypothesis and underscores the limitations of directly translating event-continuation benchmarks for less-resourced languages.

The evaluation pipeline, scoring rules, and answer extraction procedures are identical in zero- and few-shot settings; observed differences therefore arise from model outputs rather than from the evaluation protocol.

\textbf{Cross-lingual transfer and translation effects.}
{On \textit{HellaSwag}, this pattern reflects} translation-induced plausibility shifts---changes in colloquial flow, discourse markers, and event continuity that disrupt completion naturalness.
This supports the use of natively authored event-continuation benchmarks rather than literal translations.

\textbf{Dataset quality and cultural alignment.}
The native components of KyrgyzLLM-Bench, particularly \emph{KyrgyzMMLU} and \emph{KyrgyzRC}, show that LLMs trained predominantly on non-Turkic corpora exhibit partial transfer, but with accuracy substantially below English counterparts.
This highlights both data scarcity and cultural--linguistic mismatches, as idioms, syntax, and referential forms in Kyrgyz differ markedly from Indo-European patterns.
Consequently, even instruction-tuned multilingual models may misinterpret pragmatic cues and culturally grounded reasoning.
Expanding native Kyrgyz corpora and developing pretraining data with balanced linguistic registers are essential next steps.

\textbf{Actionable recommendations.}
Based on our analysis, we suggest the following improvements for future KyrgyzLLM-Bench releases and for practical use of multilingual LLMs on Kyrgyz-language tasks:
\begin{inparaenum}[(1)]
    \item enforce strict multiple-choice formatting in prompts and robust parsing for answer extraction;
    \item audit translated datasets (especially \textit{HellaSwag}) for cultural and plausibility alignment; consider fully native Kyrgyz rewrites;
    \item provide subject- and genre-level breakdowns for \emph{KyrgyzMMLU} and \emph{KyrgyzRC} to reveal domain-specific strengths and weaknesses;
    \item explore chain-of-thought prompting and rationale-based few-shot examples to test higher-order reasoning;
    \item {consider logit-based option scoring as a potential alternative to text-based parsing in order to reduce format sensitivity and improve replicability.}
\end{inparaenum}

Overall, model scaling, instruction tuning, and, in some cases, in-context learning contribute to improved performance on Kyrgyz-language tasks, yet the gap between English and Kyrgyz remains substantial. This disparity reflects imbalances in multilingual pretraining data and underscores the need for more culturally grounded evaluation and training resources.

\section{Conclusion}\label{sec:conclusion}

We present a systematic evaluation of large language models for Kyrgyz using \emph{KyrgyzLLM-Bench}, analyzing model performance under zero-shot and few-shot settings and providing a detailed account of benchmark composition, construction, and annotation.
It consists of three components:
\begin{inparaenum}[(i)]
    \item \emph{KyrgyzMMLU}, a large-scale multitask multiple-choice dataset derived from the national curriculum;
    \item \emph{KyrgyzRC}, a native reading comprehension dataset built from authentic Kyrgyz texts across encyclopedic, literary, journalistic, and mathematical domains;
    \item a translated benchmark set encompassing \textit{WinoGrande}, \textit{HellaSwag}, \textit{BoolQ}, and \textit{TruthfulQA}, enabling cross-lingual evaluation of commonsense reasoning, comprehension, and factual robustness.
\end{inparaenum}
We evaluated 26 multilingual open-source and proprietary LLMs under zero-shot and few-shot conditions, revealing substantial variability across tasks, subjects, and prompting regimes.
While modern instruction-tuned models demonstrate notable generalization capabilities, their performance on natively authored Kyrgyz tasks remains substantially below English baselines, highlighting persistent challenges in less-resourced and morphologically rich languages.

Our analysis further shows that evaluation methodology strongly influences measured outcomes, particularly for benchmarks that rely on translated prompts or culturally sensitive plausibility judgments.
Tasks such as \emph{HellaSwag} illustrate the limitations of automatic multilingual benchmarking and underscore the importance of natively authored or carefully post-edited datasets for reliable and interpretable evaluation.

Across multiple tasks, we observe that few-shot prompting can lead to unpredictable performance changes, including accuracy drops relative to zero-shot settings, especially for translated benchmarks and for models already operating near saturation. We hypothesize that this instability arises from a combination of factors: prompt translation artifacts, increased sensitivity to example ordering and surface form in morphologically rich languages, and interactions between in-context demonstrations and instruction-tuning objectives that were predominantly optimized for English or other languages. As a result, few-shot evaluation in low-resource languages should not be assumed to be uniformly beneficial and must be interpreted with caution, particularly when translated prompts or plausibility-based tasks are involved.

KyrgyzLLM-Bench fills a critical gap in the evaluation of less-resourced languages by providing culturally grounded benchmarks for Kyrgyz, an underrepresented Turkic language, and enabling systematic analysis of model behavior across native and translated tasks.
We hope this work will motivate broader inclusion of Kyrgyz in multilingual benchmarks, support more equitable progress in LLM development for Central Asian languages, and improve the accessibility of AI technologies across diverse linguistic communities.
We release datasets, evaluation code, and model results to facilitate future research and reproducibility.

\section*{Limitations}\label{subsec:limits}

Several limitations of this study should be noted.

First, performance on translated event-continuation benchmarks, most notably Kyrgyz \textit{HellaSwag}, likely underestimates attainable model capability due to translation-induced plausibility shifts. Changes in discourse flow, event sequencing, and colloquial coherence introduced during translation can substantially alter task difficulty and completion naturalness. As a result, low scores on such benchmarks should not be interpreted as definitive evidence of weak commonsense reasoning in Kyrgyz. Future work will address this limitation by developing natively authored event-continuation datasets.

Second, few-shot evaluation with prompts and exemplars translated into Kyrgyz introduces additional sources of sensitivity that complicate direct comparison with zero-shot results. Few-shot performance was observed to be highly variable across tasks and models, and in some cases degraded relative to zero-shot evaluation. This instability likely reflects interactions between translated prompt structure, example ordering, morphological complexity, and instruction-tuning objectives optimized primarily for English. We therefore recommend interpreting few-shot results in translated-prompt settings with caution and considering zero-shot performance as a complementary and more stable reference point.

Third, while multiple-choice formatting enables automatic evaluation and comparability across models, deviations from strict answer formatting can still affect measured accuracy. Enforcing stricter multiple-choice constraints in prompting and adopting alternative scoring strategies, such as logit-based option ranking, may reduce format sensitivity and improve robustness in future evaluations.

{Fourth, formal inter-annotator agreement (e.g., Cohen's $\kappa$) was not computed for \textit{KyrgyzRC}. The annotation pipeline was sequential rather than parallel: each item was authored by one student, verified by a domain supervisor, and finalized by a linguist. While this multi-stage review provides quality control, it does not yield an agreement statistic comparable to those obtained from independent dual annotation. Future versions of the dataset will incorporate a parallel-annotation phase on a held-out sample to support standard IAA reporting.}

Finally, comparisons involving proprietary models are subject to external service constraints beyond the control of this study. In particular, Gemini 2.5 Flash exhibited safety-related refusals for otherwise benign Kyrgyz-language prompts, which affected result completeness and reliability. Consequently, scores reported for such models should be interpreted with caution, as they may reflect service-level filtering behavior rather than intrinsic model capability.

{\section*{Statements and Declarations}



\paragraph{Ethics Approval and Informed Consent} The construction of \textit{KyrgyzRC} (item authoring, domain-level curation, and linguistic review; Section~\ref{subsec:kyrgyzrc}) and the translation and post-editing of \textit{WinoGrande}, \textit{HellaSwag}, \textit{BoolQ}, and \textit{TruthfulQA} (Section~\ref{subsec:translated}) were carried out by the
same cohort in an academic setting, as part of a supervised university course at the Department of Computational Linguistics. All participants---19 students and 4 supervisors/curators, all native Kyrgyz speakers---were informed in advance about the purpose of data preparation, both for natively authored and translated benchmarks, the intended research use of the datasets, and the goals of the study. Participation was voluntary and took place as part of the students' regular practical training.
The tasks were aligned with the course learning objectives and the participants' primary field of study; in consultation with university representatives, it was verified that course credit and practical experience constituted adequate compensation for the time and effort required. No sensitive personal data were collected.

\paragraph{Data and Code Availability} All datasets, evaluation code, and per-model results associated with this work are publicly released. The benchmark code is available at \url{https://github.com/golden-ratio/kyrgyzLLM_bench}. Dataset releases (the full benchmark and the \emph{KyrgyzLLM Tiny Bench} subset) are hosted on the HuggingFace Hub at \url{https://huggingface.co/collections/TTimur/kyrgyzllm-bench}. The Kyrgyz tasks have additionally been integrated into the \emph{Lighteval} evaluation framework (see \url{https://github.com/huggingface/lighteval/issues/1036}).

{\paragraph{Licensing} The KyrgyzLLM-Bench evaluation code is released under the MIT license. \textit{KyrgyzMMLU} is released under CC-BY-NC-4.0, inheriting the NonCommercial term of the GRT source materials supplied under CC-BY-NC-4.0 by the Department for Development of Education Quality; \textit{KyrgyzRC} is released under CC-BY-SA-4.0, since the Kyrgyz Wikipedia--derived subset inherits the ShareAlike obligation of CC-BY-SA-3.0 (Creative Commons designates CC-BY-SA-4.0 as a compatible later version for upgrade). The post-edited Kyrgyz translations of \textit{WinoGrande}, \textit{HellaSwag}, \textit{BoolQ}, and \textit{TruthfulQA} are released under CC-BY-4.0, MIT, CC-BY-SA-4.0, and Apache-2.0 respectively, matching their upstream licenses.}

\paragraph{Author Contributions}%
Conceptualization, Data curation, Resources, Project administration: T.~Turatali. Data curation, Investigation, Validation: A.~Turdubaeva. Investigation, Software, Validation: R.~Izmailov. Methodology, Formal analysis, Writing --- original draft, review \& editing: A.~Alekseev. Formal analysis, Writing --- review \& editing: S.~Nikolenko. All authors read and approved the final manuscript.

\paragraph{Acknowledgements}
The work of S. I. Nikolenko was supported by the Ministry of Science and Higher Education of the Russian Federation (agreement 075-15-2025-344 dated 29/04/2025 for Saint Petersburg Leonhard Euler International Mathematical Institute at PDMI RAS).

\bibliographystyle{abbrv}
\bibliography{references}

@inproceedings{lai2023okapi,
    author={Lai, Viet Dac and Van Nguyen, Chien and Ngo, Nghia Trung and Nguyen, Thuat and Dernoncourt, Franck and Rossi, Ryan A. and others},
    booktitle={Proceedings of the 2023 Conference on Empirical Methods in Natural Language Processing: System Demonstrations},
    title={Okapi: Instruction-tuned large language models in multiple languages with reinforcement learning from human feedback},
    year={2023},
    pages={318--327}
}

@INPROCEEDINGS{vanmassenhove2021,
    author={Vanmassenhove, E. and Shterionov, D. and Gwilliam, M.},
    booktitle={Proceedings of the 16th Conference of the European Chapter of the Association for Computational Linguistics: Main Volume},
    title={Machine Translationese: Effects of Algorithmic Bias on Linguistic Complexity in Machine Translation},
    year={2021},
    pages={2203--2213}
}

@INPROCEEDINGS{wang2018glue,
    author={Wang, Alex and Singh, Amanpreet and Michael, Julian and Hill, Felix and Levy, Omer and Bowman, Samuel R.},
    booktitle={Proceedings of the 2018 EMNLP Workshop BlackboxNLP},
    title={GLUE: A multi-task benchmark and analysis platform for natural language understanding},
    year={2018},
    pages={353--355}
}

@inproceedings{wang2019superglue,
    author={Wang, Alex and Pruksachatkun, Yada and Nangia, Nikita and Singh, Amanpreet and Michael, Julian and Hill, Felix},
    booktitle={Advances in Neural Information Processing Systems},
    title={SuperGLUE: A stickier benchmark for general-purpose language understanding systems},
    year={2019}
}

@article{hendrycks2021measuring,
  title={Measuring Massive Multitask Language Understanding},
  author={Hendrycks, Dan and Burns, Collin and  Basart, Steven and  Zou, Andy and  Mazeika, Mantas and  Song, Dawn and  Steinhardt, Jacob},
  journal={Proceedings of the International Conference on Learning Representations (ICLR)},
  year={2021}
}

@inproceedings{hu2020xtreme,
    title={Xtreme: A massively multilingual multi-task benchmark for evaluating cross-lingual generalization},
    author={Hu, Junjie and Ruder, Sebastian and Siddhant, Aditya and Neubig, Graham and Firat, Orhan and Johnson, Melvin},
    booktitle={Proceedings of the 37th International Conference on Machine Learning},
    volume={119},
    pages={4411--4421},
    year={2020}
}

@INPROCEEDINGS{goyal2022flores,
    author={Goyal, Naman and Gao, Cynthia and Chaudhary, Vishrav and Chen, Peng-Jen and Wenzek, Guillaume and Ju, Da and others},
    booktitle={Proceedings of ACL},
    title={The FLORES-101 evaluation benchmark for low-resource and multilingual machine translation},
    year={2022}
}

@inproceedings{skadina2025evaluation,
  author    = {Skadi{\c{n}}a, Inguna and Bakanovs, Bruno and Dar{\c{g}}is, Roberts},
  title     = {First Steps in Benchmarking {L}atvian in Large Language Models},
  booktitle = {Proceedings of the Third Workshop on Resources and Representations
               for Under-Resourced Languages and Domains (RESOURCEFUL-2025)},
  publisher = {University of Tartu Library},
  year      = {2025},
  pages     = {86--95},
  url       = {https://hdl.handle.net/10062/107120}
}

@INPROCEEDINGS{dargis2024evaluating,
    author={Dar{\c{g}}is, Roberts and B{\={a}}rzdi{\c{n}}{\v{s}}, Guntis and Skadi{\c{n}}a, Inguna and Gr{\={u}}z{\={\i}}tis, Normunds and Saul{\={\i}}te, Baiba},
    booktitle={Proceedings of the 4th International Conference on Natural Language Processing for Digital Humanities},
    title={Evaluating Open-Source LLMs in Low-Resource Languages: Insights from Latvian High School Exams},
    year={2024},
    pages={289--293}
}

@INPROCEEDINGS{zellers2019hellaswag_ieee,
    author={Zellers, Rowan and Holtzman, Ari and Bisk, Yonatan and Farhadi, Ali and Choi, Yejin},
    booktitle={Proc. Conf. Empirical Methods in Natural Language Processing (EMNLP)},
    title={HellaSwag: Can a Machine Really Finish Your Sentence?},
    year={2019},
    address={Hong Kong, China},
    month={Nov.}
}

@INPROCEEDINGS{sakaguchi2020winogrande_ieee,
    author={Sakaguchi, Keisuke and Bras, Ronan Le and Bhagavatula, Chandra and Choi, Yejin},
    booktitle={Proc. Thirty-Fourth AAAI Conf. on Artificial Intelligence (AAAI)},
    title={WinoGrande: An Adversarial Winograd Schema Challenge at Scale},
    year={2020},
    address={New York, NY, USA},
    month={Feb.}
}

@INPROCEEDINGS{clark2019boolq,
    author={Clark, Christopher and Lee, Kenton and Chang, Ming-Wei and Kwiatkowski, Tom and Collins, Michael and Toutanova, Kristina},
    booktitle={Proc. Annual Conference of the North American Chapter of the Association for Computational Linguistics (NAACL)},
    title={BoolQ: Exploring the Surprising Difficulty of Natural Yes/No Questions},
    year={2019},
    address={Minneapolis, MN, USA},
    month={Jun.}
}

@INPROCEEDINGS{cobbe2021training_ieee,
    author={Cobbe, Karl and Kosaraju, Vineet and Bavarian, Mohammad and Chen, Mark and Jun, Heewoo and Kaiser, Lukasz and others},
    booktitle={Proc. International Conference on Learning Representations (ICLR)},
    title={Training Verifiers to Solve Math Word Problems},
    year={2021},
    address={Virtual},
    month={May}
}

@INPROCEEDINGS{lin2022truthfulqa,
    author={Lin, Z. and Hilton, J. and Evans, O.},
    booktitle={Proc. Annual Meeting of the Association for Computational Linguistics (ACL)},
    title={TruthfulQA: Measuring How Models Mimic Human Falsehoods},
    year={2022},
    address={Dublin, Ireland},
    month={May}
}

@inproceedings{alekseev2024kyrgyznlp,
  title={{KyrgyzNLP}: challenges, progress, and future},
  author={Alekseev, Anton and Turatali, Timur},
  booktitle={International Conference on Analysis of Images, Social Networks and Texts},
  pages={3--39},
  year={2024},
  organization={Springer}
}

@MISC{akylai_24kg,
    title = {AkylAI smart speaker: Artificial intelligence speaking Kyrgyz language (June 18th, 2024)},
    author = {Kan, Anastasia},
    year = {2024},
    howpublished = {\url{https://web.archive.org/web/20240619010036/https://24.kg/english/296874_AkylAI_smart_speaker_Artificial_intelligence_speaking_Kyrgyz_language/}},
    note = {Accessed: 2024-09-14}
}

@MISC{sexed_kyrgyz,
    title = {A chatbot for teenagers about puberty, relationships, and health launched in Kyrgyzstan (May 24th, 2022)},
    author = {UNESCO-IITE},
    year = {2022},
    howpublished = {\url{http://web.archive.org/web/20240525072322/https://iite.unesco.org/highlights/oilo-chatbot-sex-ed-kyrgyzstan-en/}},
    note = {Accessed: 2024-09-14}
}

@INPROCEEDINGS{mirzakhalov2021evaluating,
    title={Evaluating Multiway Multilingual NMT in the Turkic Languages},
    author={Mirzakhalov, Jamshidbek and Babu, Anoop and Kunafin, Aigiz and Wahab, Ahsan and Moydinboyev, Behzod and Ivanova, Sardana and others},
    booktitle={Proceedings of the Sixth Conference on Machine Translation},
    pages={518--530},
    year={2021}
}

@inproceedings{veitsman2025,
  title={Recent Advancements and Challenges of Turkic Central Asian Language Processing},
  author={Veitsman, Yana and Hartmann, Mareike},
  booktitle={Proceedings of the First Workshop on Language Models for Low-Resource Languages},
  pages={309--324},
  year={2025}
}

@misc{lighteval,
    author = {Habib, Nathan and Fourrier, Clémentine and Kydlíček, Hynek and Wolf, Thomas and Tunstall, Lewis},
    title = {LightEval: A lightweight framework for LLM evaluation},
    year = {2023},
    version = {0.11.0},
    url = {https://github.com/huggingface/lighteval}
}

@misc{grattafiori2024llama3herdmodels,
      title={The Llama 3 Herd of Models}, 
      author={Grattafiori, Aaron and Dubey, Abhimanyu and Jauhri, Abhinav and Pandey, Abhinav and Kadian, Abhishek and Al-Dahle, Ahmad and Letman, Aiesha and Mathur, Akhil and Schelten, Alan and Vaughan, Alex and Yang, Amy and Fan, Angela and Goyal, Anirudh and Hartshorn, Anthony and Yang, Aobo and Mitra, Archi and Sravankumar, Archie and Korenev, Artem and Hinsvark, Arthur and Rao, Arun and Zhang, Aston and Rodriguez, Aurelien and Gregerson, Austen and Spataru, Ava and Roziere, Baptiste and Biron, Bethany and Tang, Binh and Chern, Bobbie and Caucheteux, Charlotte and Nayak, Chaya and Bi, Chloe and Marra, Chris and McConnell, Chris and Keller, Christian and Touret, Christophe and Wu, Chunyang and Wong, Corinne and Canton Ferrer, Cristian and Nikolaidis, Cyrus and Allonsius, Damien and Song, Daniel and Pintz, Danielle and Livshits, Danny and Wyatt, Danny and Esiobu, David and Choudhary, Dhruv and Mahajan, Dhruv and Garcia-Olano, Diego and Perino, Diego and Hupkes, Dieuwke and Lakomkin, Egor and AlBadawy, Ehab and Lobanova, Elina and Dinan, Emily and Smith, Eric Michael and Radenovic, Filip and Guzmán, Francisco and Zhang, Frank and Synnaeve, Gabriel and Lee, Gabrielle and Lewis Anderson, Georgia and Thattai, Govind and Nail, Graeme and Mialon, Gregoire and Pang, Guan and Cucurell, Guillem and Nguyen, Hailey and Korevaar, Hannah and Xu, Hu and Touvron, Hugo and Zarov, Iliyan and Arrieta Ibarra, Imanol and Kloumann, Isabel and Misra, Ishan and Evtimov, Ivan and Zhang, Jack and Copet, Jade and Lee, Jaewon and Geffert, Jan and Vranes, Jana and Park, Jason and Mahadeokar, Jay and Shah, Jeet and van der Linde, Jelmer and Billock, Jennifer and Hong, Jenny and Lee, Jenya and Fu, Jeremy and Chi, Jianfeng and Huang, Jianyu and Liu, Jiawen and Wang, Jie and Yu, Jiecao and Bitton, Joanna and Spisak, Joe and Park, Jongsoo and Rocca, Joseph and Johnstun, Joshua and Saxe, Joshua and Jia, Junteng and Alwala, Kalyan Vasuden and Prasad, Karthik and Upasani, Kartikeya and Plawiak, Kate and Li, Ke and Heafield, Kenneth and Stone, Kevin and El-Arini, Khalid and Iyer, Krithika and Malik, Kshitiz and Chiu, Kuenley and Bhalla, Kunal and Lakhotia, Kushal and Rantala-Yeary, Lauren and van der Maaten, Laurens and Chen, Lawrence and Tan, Liang and Jenkins, Liz and Martin, Louis and Madaan, Lovish and Malo, Lubo and Blecher, Lukas and Landzaat, Lukas and de Oliveira, Luke and Muzzi, Madeline and Pasupuleti, Mahesh and Singh, Mannat and Paluri, Manohar and Kardas, Marcin and Tsimpoukelli, Maria and Oldham, Mathew and Rita, Mathieu and Pavlova, Maya and Kambadur, Melanie and Lewis, Mike and Si, Min and Kumar Singh, Mitesh and Hassan, Mona and Goyal, Naman and Torabi, Narjes and Bashlykov, Nikolay and Bogoychev, Nikolay and Chatterji, Niladri and Zhang, Ning and Duchenne, Olivier and Çelebi, Onur and Alrassy, Patrick and Zhang, Pengchuan and Li, Pengwei and Vasic, Petar and Weng, Peter and Bhargava, Prajjwal and Dubal, Pratik and Krishnan, Praveen and Singh Koura, Punit and Xu, Puxin and He, Qing and Dong, Qingxiao and Srinivasan, Ragavan and Ganapathy, Raj and Calderer, Ramon and Silveira Cabral, Ricardo and Stojnic, Robert and Raileanu, Roberta and Maheswari, Rohan and Girdhar, Rohit and Patel, Rohit and Sauvestre, Romain and Polidoro, Ronnie and Sumbaly, Roshan and Taylor, Ross and Silva, Ruan and Hou, Rui and Wang, Rui and Hosseini, Saghar and Chennabasappa, Sahana and Singh, Sanjay and Bell, Sean and Kim, Seohyun Sonia and Edunov, Sergey and Nie, Shaoliang and Narang, Sharan and Raparthy, Sharath and Shen, Sheng and Wan, Shengye and Bhosale, Shruti and Zhang, Shun and Vandenhende, Simon and Batra, Soumya and Whitman, Spencer and Sootla, Sten and Collot, Stephane and Gururangan, Suchin and Borodinsky, Sydney and Herman, Tamar and Fowler, Tara and Sheasha, Tarek and Georgiou, Thomas and Scialom, Thomas and Speckbacher, Tobias and Mihaylov, Todor and Xiao, Tong and Karn, Ujjwal and Goswami, Vedanuj and Gupta, Vibhor and Ramanathan, Vignesh and Kerkez, Viktor and Gonguet, Vincent and Do, Virginie and Vogeti, Vish and Albiero, Vítor and Petrovic, Vladan and Chu, Weiwei and Xiong, Wenhan and Fu, Wenyin and Meers, Whitney and Martinet, Xavier and Wang, Xiaodong and Wang, Xiaofang and Tan, Xiaoqing Ellen and Xia, Xide and Xie, Xinfeng and Jia, Xuchao and Wang, Xuewei and Goldschlag, Yaelle and Gaur, Yashesh and Babaei, Yasmine and Wen, Yi and Song, Yiwen and Zhang, Yuchen and Li, Yue and Mao, Yuning and Delpierre Coudert, Zacharie and Yan, Zheng and Chen, Zhengxing and Papakipos, Zoe and Singh, Aaditya and Srivastava, Aayushi and Jain, Abha and Kelsey, Adam and Shajnfeld, Adam and Gangidi, Adithya and Victoria, Adolfo and Goldstand, Ahuva and Menon, Ajay and Sharma, Ajay and Boesenberg, Alex and Baevski, Alexei and Feinstein, Allie and Kallet, Amanda and Sangani, Amit and Teo, Amos and Yunus, Anam and Lupu, Andrei and Alvarado, Andres and Caples, Andrew and Gu, Andrew and Ho, Andrew and Poulton, Andrew and Ryan, Andrew and Ramchandani, Ankit and Dong, Annie and Franco, Annie and Goyal, Anuj and Saraf, Aparajita and Chowdhury, Arkabandhu and Gabriel, Ashley and Bharambe, Ashwin and Eisenman, Assaf and Yazdan, Azadeh and James, Beau and Maurer, Ben and Leonhardi, Benjamin and Huang, Bernie and Loyd, Beth and De Paola, Beto and Paranjape, Bhargavi and Liu, Bing and Wu, Bo and Ni, Boyu and Hancock, Braden and Wasti, Bram and Spence, Brandon and Stojkovic, Brani and Gamido, Brian and Montalvo, Britt and Parker, Carl and Burton, Carly and Mejia, Catalina and Liu, Ce and Wang, Changhan and Kim, Changkyu and Zhou, Chao and Hu, Chester and Chu, Ching-Hsiang and Cai, Chris and Tindal, Chris and Feichtenhofer, Christoph and Gao, Cynthia and Civin, Damon and Beaty, Dana and Kreymer, Daniel and Li, Daniel and Adkins, David and Xu, David and Testuggine, Davide and David, Delia and Parikh, Devi and Liskovich, Diana and Foss, Didem and Wang, Dingkang and Le, Duc and Holland, Dustin and Dowling, Edward and Jamil, Eissa and Montgomery, Elaine and Presani, Eleonora and Hahn, Emily and Wood, Emily and Le, Eric-Tuan and Brinkman, Erik and Arcaute, Esteban and Dunbar, Evan and Smothers, Evan and Sun, Fei and Kreuk, Felix and Tian, Feng and Kokkinos, Filippos and Ozgenel, Firat and Caggioni, Francesco and Kanayet, Frank and Seide, Frank and Medina Florez, Gabriela and Schwarz, Gabriella and Badeer, Gada and Swee, Georgia and Halpern, Gil and Herman, Grant and Sizov, Grigory and Zhang, Guangyi and Lakshminarayanan, Guna and Inan, Hakan and Shojanazeri, Hamid and Zou, Han and Wang, Hannah and Zha, Hanwen and Habeeb, Haroun and Rudolph, Harrison and Suk, Helen and Aspegren, Henry and Goldman, Hunter and Zhan, Hongyuan and Damlaj, Ibrahim and Molybog, Igor and Tufanov, Igor and Leontiadis, Ilias and Veliche, Irina-Elena and Gat, Itai and Weissman, Jake and Geboski, James and Kohli, James and Lam, Janice and Asher, Japhet and Gaya, Jean-Baptiste and Marcus, Jeff and Tang, Jeff and Chan, Jennifer and Zhen, Jenny and Reizenstein, Jeremy and Teboul, Jeremy and Zhong, Jessica and Jin, Jian and Yang, Jingyi and Cummings, Joe and Carvill, Jon and Shepard, Jon and McPhie, Jonathan and Torres, Jonathan and Ginsburg, Josh and Wang, Junjie and Wu, Kai and U, Kam Hou and Saxena, Karan and Khandelwal, Kartikay and Zand, Katayoun and Matosich, Kathy and Veeraraghavan, Kaushik and Michelena, Kelly and Li, Keqian and Jagadeesh, Kiran and Huang, Kun and Chawla, Kunal and Huang, Kyle and Chen, Lailin and Garg, Lakshya and A, Lavender and Silva, Leandro and Bell, Lee and Zhang, Lei and Guo, Liangpeng and Yu, Licheng and Moshkovich, Liron and Wehrstedt, Luca and Khabsa, Madian and Avalani, Manav and Bhatt, Manish and Mankus, Martynas and Hasson, Matan and Lennie, Matthew and Reso, Matthias and Groshev, Maxim and Naumov, Maxim and Lathi, Maya and Keneally, Meghan and Liu, Miao and Seltzer, Michael L. and Valko, Michal and Restrepo, Michelle and Patel, Mihir and Vyatskov, Mik and Samvelyan, Mikayel and Clark, Mike and Macey, Mike and Wang, Mike and Jubert Hermoso, Miquel and Metanat, Mo and Rastegari, Mohammad and Bansal, Munish and Santhanam, Nandhini and Parks, Natascha and White, Natasha and Bawa, Navyata and Singhal, Nayan and Egebo, Nick and Usunier, Nicolas and Mehta, Nikhil and Pavlovich Laptev, Nikolay and Dong, Ning and Cheng, Norman and Chernoguz, Oleg and Hart, Olivia and Salpekar, Omkar and Kalinli, Ozlem and Kent, Parkin and Parekh, Parth and Saab, Paul and Balaji, Pavan and Rittner, Pedro and Bontrager, Philip and Roux, Pierre and Dollar, Piotr and Zvyagina, Polina and Ratanchandani, Prashant and Yuvraj, Pritish and Liang, Qian and Alao, Rachad and Rodriguez, Rachel and Ayub, Rafi and Murthy, Raghotham and Nayani, Raghu and Mitra, Rahul and Parthasarathy, Rangaprabhu and Li, Raymond and Hogan, Rebekkah and Battey, Robin and Wang, Rocky and Howes, Russ and Rinott, Ruty and Mehta, Sachin and Siby, Sachin and Bondu, Sai Jayesh and Datta, Samyak and Chugh, Sara and Hunt, Sara and Dhillon, Sargun and Sidorov, Sasha and Pan, Satadru and Mahajan, Saurabh and Verma, Saurabh and Yamamoto, Seiji and Ramaswamy, Sharadh and Lindsay, Shaun and Lindsay, Shaun and Feng, Sheng and Lin, Shenghao and Cindy Zha, Shengxin and Patil, Shishir and Shankar, Shiva and Zhang, Shuqiang and Zhang, Shuqiang and Wang, Sinong and Agarwal, Sneha and Sajuyigbe, Soji and Chintala, Soumith and Max, Stephanie and Chen, Stephen and Kehoe, Steve and Satterfield, Steve and Govindaprasad, Sudarshan and Gupta, Sumit and Deng, Summer and Cho, Sungmin and Virk, Sunny and Subramanian, Suraj and Choudhury, Sy and Goldman, Sydney and Remez, Tal and Glaser, Tamar and Best, Tamara and Koehler, Thilo and Robinson, Thomas and Li, Tianhe and Zhang, Tianjun and Matthews, Tim and Chou, Timothy and Shaked, Tzook and Vontimitta, Varun and Ajayi, Victoria and Montanez, Victoria and Mohan, Vijai and Satish Kumar, Vinay and Mangla, Vishal and Ionescu, Vlad and Poenaru, Vlad and Tiberiu Mihailescu, Vlad and Ivanov, Vladimir and Li, Wei and Wang, Wenchen and Jiang, Wenwen and Bouaziz, Wes and Constable, Will and Tang, Xiaocheng and Wu, Xiaojian and Wang, Xiaolan and Wu, Xilun and Gao, Xinbo and Kleinman, Yaniv and Chen, Yanjun and Hu, Ye and Jia, Ye and Qi, Ye and Li, Yenda and Zhang, Yilin and Zhang, Ying and Adi, Yossi and Nam, Youngjin and Wang, Yu and Zhao, Yu and Hao, Yuchen and Qian, Yundi and Li, Yunlu and He, Yuzi and Rait, Zach and DeVito, Zachary and Rosnbrick, Zef and Wen, Zhaoduo and Yang, Zhenyu and Zhao, Zhiwei and Ma, Zhiyu},
      year={2024},
      eprint={2407.21783},
      archivePrefix={arXiv},
      primaryClass={cs.AI},
      url={https://arxiv.org/abs/2407.21783}, 
}

@misc{gemmateam2024gemmaopenmodelsbased,
      title={Gemma: Open Models Based on Gemini Research and Technology}, 
      author={GemmaTeam and Mesnard, Thomas and Hardin, Cassidy and Dadashi, Robert and Bhupatiraju, Surya and Pathak, Shreya and Sifre, Laurent and Rivière, Morgane and Kale, Mihir Sanjay and Love, Juliette and Tafti, Pouya and Hussenot, Léonard and Sessa, Pier Giuseppe and Chowdhery, Aakanksha and Roberts, Adam and Barua, Aditya and Botev, Alex and Castro-Ros, Alex and Slone, Ambrose and Héliou, Amélie and Tacchetti, Andrea and Bulanova, Anna and Paterson, Antonia and Tsai, Beth and Shahriari, Bobak and Le Lan, Charline and Choquette-Choo, Christopher A. and Crepy, Clément and Cer, Daniel and Ippolito, Daphne and Reid, David and Buchatskaya, Elena and Ni, Eric and Noland, Eric and Yan, Geng and Tucker, George and Muraru, George-Christian and Rozhdestvenskiy, Grigory and Michalewski, Henryk and Tenney, Ian and Grishchenko, Ivan and Austin, Jacob and Keeling, James and Labanowski, Jane and Lespiau, Jean-Baptiste and Stanway, Jeff and Brennan, Jenny and Chen, Jeremy and Ferret, Johan and Chiu, Justin and Mao-Jones, Justin and Lee, Katherine and Yu, Kathy and Millican, Katie and Lowe Sjoesund, Lars and Lee, Lisa and Dixon, Lucas and Reid, Machel and Mikuła, Maciej and Wirth, Mateo and Sharman, Michael and Chinaev, Nikolai and Thain, Nithum and Bachem, Olivier and Chang, Oscar and Wahltinez, Oscar and Bailey, Paige and Michel, Paul and Yotov, Petko and Chaabouni, Rahma and Comanescu, Ramona and Jana, Reena and Anil, Rohan and McIlroy, Ross and Liu, Ruibo and Mullins, Ryan and Smith, Samuel L and Borgeaud, Sebastian and Girgin, Sertan and Douglas, Sholto and Pandya, Shree and Shakeri, Siamak and De, Soham and Klimenko, Ted and Hennigan, Tom and Feinberg, Vlad and Stokowiec, Wojciech and Chen, Yu-hui and Ahmed, Zafarali and Gong, Zhitao and Warkentin, Tris and Peran, Ludovic and Giang, Minh and Farabet, Clément and Vinyals, Oriol and Dean, Jeff and Kavukcuoglu, Koray and Hassabis, Demis and Ghahramani, Zoubin and Eck, Douglas and Barral, Joelle and Pereira, Fernando and Collins, Eli and Joulin, Armand and Fiedel, Noah and Senter, Evan and Andreev, Alek and Kenealy, Kathleen
},
      year={2024},
      eprint={2403.08295},
      archivePrefix={arXiv},
      primaryClass={cs.CL},
      url={https://arxiv.org/abs/2403.08295}, 
}

@misc{bai2023qwentechnicalreport,
      title={Qwen Technical Report}, 
      author={Bai, Jinze and Bai, Shuai and Chu, Yunfei and Cui, Zeyu and Dang, Kai and Deng, Xiaodong and Fan, Yang and Ge, Wenbin and Han, Yu and Huang, Fei and Hui, Binyuan and Ji, Luo and Li, Mei and Lin, Junyang and Lin, Runji and Liu, Dayiheng and Liu, Gao and Lu, Chengqiang and Lu, Keming and Ma, Jianxin and Men, Rui and Ren, Xingzhang and Ren, Xuancheng and Tan, Chuanqi and Tan, Sinan and Tu, Jianhong and Wang, Peng and Wang, Shijie and Wang, Wei and Wu, Shengguang and Xu, Benfeng and Xu, Jin and Yang, An and Yang, Hao and Yang, Jian and Yang, Shusheng and Yao, Yang and Yu, Bowen and Yuan, Hongyi and Yuan, Zheng and Zhang, Jianwei and Zhang, Xingxuan and Zhang, Yichang and Zhang, Zhenru and Zhou, Chang and Zhou, Jingren and Zhou, Xiaohuan and Zhu, Tianhang},
      year={2023},
      eprint={2309.16609},
      archivePrefix={arXiv},
      primaryClass={cs.CL},
      url={https://arxiv.org/abs/2309.16609}, 
}

@INPROCEEDINGS{turatali2025bridging,
  author={Turatali, Timur and Turdubaeva, Aida and Zhenishbekov, Islam and Suranbaev, Zhoomart and Alekseev, Anton and Izmailov, Rustem},
  booktitle={2025 10th International Conference on Computer Science and Engineering (UBMK)}, 
  title={Bridging the Gap in Less-Resourced Languages: Building a Benchmark for Kyrgyz Language Models}, 
  year={2025},
  volume={},
  number={},
  pages={1673-1677},
  doi={10.1109/UBMK67458.2025.11206960}
}

@article{wu2025bitter,
  title={The bitter lesson learned from 2,000+ multilingual benchmarks},
  author={Wu, Minghao and Wang, Weixuan and Liu, Sinuo and Yin, Huifeng and Wang, Xintong and Zhao, Yu and Lyu, Chenyang and Wang, Longyue and Luo, Weihua and Zhang, Kaifu},
  journal={arXiv preprint arXiv:2504.15521},
  year={2025}
}

@article{Salmorbekova2023KyrgyzRussian,
  author  = {Salmorbekova, R. and Alymbaev, A. and Tukhtamatov, A.},
  title   = {Kyrgyz and Russian Languages in the Eurasian Space},
  journal = {Bulletin of Science and Practice},
  year    = {2023},
  volume  = {9},
  number  = {6},
  pages   = {722--733},
  doi     = {10.33619/2414-2948/91/93},
  note    = {In Russian},
}

@inproceedings{jumashev2025kyrgyz,
  title={The Kyrgyz Seed Dataset Submission to the WMT25 Open Language Data Initiative Shared Task},
  author={Jumashev, Murat and Tillabaeva, Alina and Kasieva, Aida and Omurkanov, Turgunbek and Musaeva, Akylai and Kyzy, Meerim Emil and Chagataeva, Gulaiym and Washington, Jonathan},
  booktitle={Proceedings of the Tenth Conference on Machine Translation},
  pages={1088--1102},
  year={2025}
}

@misc{hf_open_llm_leaderboard,
  title        = {Open LLM Leaderboard (Archived Evaluation Protocol)},
  author       = {{Hugging Face}},
  howpublished = {\url{https://huggingface.co/docs/leaderboards/en/open_llm_leaderboard/archive}},
  year         = {2024}
}

@misc{lm_eval_harness,
  title        = {Language Model Evaluation Harness},
  author       = {Gao, Leo and Tow, Jonathan and Biderman, Stella and Black, Sid and DiPofi, Albert and Foster, Charles and Golding, Laurence and Hsu, Jeffrey and McDonell, Kyle and Muennighoff, Niklas and others},
  howpublished = {\url{https://github.com/EleutherAI/lm-evaluation-harness}},
  year         = {2023}
}

@misc{nemo_inference_guidelines,
  title        = {NVIDIA NeMo Microservices: Model Parameter Tuning Guidelines},
  author       = {{NVIDIA}},
  howpublished = {\url{https://docs.nvidia.com/nemo/microservices/latest/design-synthetic-data-from-scratch-or-seeds/configure-models.html}},
  year         = {2025}
}

@inproceedings{singh-etal-2025-global,
    title = "Global {MMLU}: Understanding and Addressing Cultural and Linguistic Biases in Multilingual Evaluation",
    author = "Singh, Shivalika  and
      Romanou, Angelika  and
      Fourrier, Cl{\'e}mentine  and
      Adelani, David Ifeoluwa  and
      Ngui, Jian Gang  and
      Vila-Suero, Daniel  and
      Limkonchotiwat, Peerat  and
      Marchisio, Kelly  and
      Leong, Wei Qi  and
      Susanto, Yosephine  and
      Ng, Raymond  and
      Longpre, Shayne  and
      Ruder, Sebastian  and
      Ko, Wei-Yin  and
      Bosselut, Antoine  and
      Oh, Alice  and
      Martins, Andre  and
      Choshen, Leshem  and
      Ippolito, Daphne  and
      Ferrante, Enzo  and
      Fadaee, Marzieh  and
      Ermis, Beyza  and
      Hooker, Sara",
    editor = "Che, Wanxiang  and
      Nabende, Joyce  and
      Shutova, Ekaterina  and
      Pilehvar, Mohammad Taher",
    booktitle = "Proceedings of the 63rd Annual Meeting of the Association for Computational Linguistics (Volume 1: Long Papers)",
    month = jul,
    year = "2025",
    address = "Vienna, Austria",
    publisher = "Association for Computational Linguistics",
    url = "https://aclanthology.org/2025.acl-long.919/",
    doi = "10.18653/v1/2025.acl-long.919",
    pages = "18761--18799",
    ISBN = "979-8-89176-251-0"
}
\clearpage
\appendix

\setcounter{listing}{0}
\renewcommand{\thelisting}{A\arabic{listing}}

\section{Prompting strategies}

{For benchmarks other than \emph{KyrgyzMMLU} or \emph{KyrgyzRC}, the prompts we have used are direct translations of original English queries to Kyrgyz; in this section, we provide the prompts for original datasets only in Listings~\ref{lst:few-prompt-rc}, \ref{lst:few-prompt-mmlu}, and \ref{lst:zero-prompt}. The translated prompts, as well as those presented here are available in the code of \textbf{KyrgyzLLM-Bench}.}

    \begin{listing}[ht]
      \caption{Prompt for few-shot solution for \textit{KyrgyzRC} (actual prompt is built dynamically in Python code, some details have been removed).}
      \label{lst:few-prompt-rc}
      \selectlanguage{russian}
      \begin{Verbatim}[fontsize=\small, frame=lines, breaklines, samepage=true]
Сизге бир темага байланыштуу бир нече үзүндү текст берилген. Бардык үзүндүлөрдү кунт коюп окуп, андан кийин төмөндөгү суроолорго жооп бериңиздер.
Суроо менен 2-4 жооп варианты берилет, туура жооптун НОМЕРИН (индексин) гана кайтарышыңыз керек.

Текст: {example_01_text}
Суроо: {example_01_question}
Сунушталган жооптор:
0. {example_01_choices[0]}
1. {example_01_choices[1]}
2. {example_01_choices[2]}
Туура жоопту тандаңыз:  {example_01_answer}

Текст: {example_02_text}
Суроо: {example_02_question}
Сунушталган жооптор:
0. {example_02_choices[0]}
1. {example_02_choices[1]}
2. {example_02_choices[2]}
3. {example_02_choices[3]}
Туура жоопту тандаңыз:  {example_02_answer}

Текст: {example_03_text}
Суроо: {example_03_question}
Сунушталган жооптор:
0. {example_03_choices[0]}
1. {example_03_choices[1]}
Туура жоопту тандаңыз:  {example_03_answer}

Текст: {text}
Суроо: {question}
Сунушталган жооптор:
0. {choices[0]}
1. {choices[1]}
2. {choices[2]}
3. {choices[3]}
Туура жоопту тандаңыз:
      \end{Verbatim}
      \selectlanguage{english}
    \end{listing}
    \begin{listing}[ht]
      \caption{Prompt for few-shot solution for \textit{KyrgyzMMLU} (actual prompt is built dynamically in Python code, some details have been removed).}
      \label{lst:few-prompt-mmlu}
      \selectlanguage{russian}
      \begin{Verbatim}[fontsize=\small, frame=lines, breaklines, samepage=true]
Сиз билимиңизге жана жөндөмүңүзгө жараша суроолорго жооп берген AIсыз.
Сизге суроо жана 2-5 жооп варианты берилет, туура жооптун НОМЕРИН (индексин) гана кайтарышыңыз керек.

Суроо: {example_01_question}
Сунушталган жооптор:
0. {example_01_choices[0]}
1. {example_01_choices[1]}
2. {example_01_choices[2]}
3. {example_01_choices[3]}
4. {example_01_choices[4]}
Туура жоопту тандаңыз:  {example_01_answer}

Суроо: {example_02_question}
Сунушталган жооптор:
0. {example_02_choices[0]}
1. {example_02_choices[1]}
2. {example_02_choices[2]}
3. {example_02_choices[3]}
Туура жоопту тандаңыз:  {example_02_answer}

Суроо: {question}
Сунушталган жооптор:
0. {choices[0]}
1. {choices[1]}
2. {choices[2]}
3. {choices[3]}
4. {choices[4]}
Туура жоопту тандаңыз:
      \end{Verbatim}
      \selectlanguage{english}
    \end{listing}
    \begin{listing}[!ht]
      \caption{Prompt for zero-shot solution for \emph{KyrgyzMMLU} / \emph{KyrgyzRC} (actual prompt is built dynamically in Python code, some details such as choices' list building have been removed for better readability).}

      \selectlanguage{russian}
      \begin{Verbatim}[frame=lines, breaklines, samepage=true]
Сиз билимиңизге жана жөндөмүңүзгө жараша суроолорго жооп берген AIсыз.
Сизге суроо жана 2-5 жооп варианты берилет, туура жооптун НОМЕРИН (индексин) гана кайтарышыңыз керек.

Текст: {text}
Суроо: {question}
Сунушталган жооптор:
  а. {choices[0]}
  б. {choices[1]}
  в. {choices[2]}
  г. {choices[3]}
Туура жоопту тандаңыз:
      \end{Verbatim}
      \selectlanguage{english}
      \label{lst:zero-prompt}
    \end{listing}

\section{\emph{GSM8K} Translation}\label{appsec:gsm8k}

Although the \emph{GSM8K} dataset was translated into Kyrgyz, we do not include its results in the main reported version of the benchmark. This decision was made for several reasons. First, the evaluation protocol of \emph{GSM8K} differs substantially from that of the other benchmarks considered in this work, as it relies on a strict exact-match metric rather than standard accuracy. Second, our preliminary experiments indicate that the standard \textit{Lighteval} evaluation script, which we adopt to ensure comparability with existing language benchmarks, is highly sensitive to output formatting. In the Kyrgyz setting, this sensitivity makes it difficult to disentangle genuine mathematical problem-solving ability from formatting effects, such as mismatches between the expected output patterns and language-specific conventions for expressing numerical values. As a result, performance on \emph{GSM8K} under this setup may reflect evaluation artifacts rather than model competence, and we therefore exclude it from the primary analysis.

More generally, this design choice reflects our intention to keep the benchmark focused on a coherent and interpretable set of evaluation signals. By restricting the reported results to benchmarks that share a common evaluation paradigm, we aim to ensure that the aggregate metrics convey a clear and atomic message about model performance in the Kyrgyz setting, without mixing heterogeneous sources of error (where possible). At the same time, we recognize that excluding numerically intensive reasoning tasks such as \emph{GSM8K} limits the scope of the current analysis. Developing evaluation protocols that can robustly accommodate language-specific numerical expressions and disentangle reasoning ability from formatting effects remains an important direction for future work.

\section{KyrgyzRC sample data entry}\label{appsec:kyrgyzrc-entry}

Each \textit{KyrgyzRC} item follows the schema described in Section~\ref{sec:benchmark}.
A representative entry, in JSON-like form, is shown below; the passage and question are
drawn from the encyclopedic (Wikipedia) subset.

\begin{verbatim}
{
  "source_type":   "wikipedia",
  "question_type": "factual_understanding",
  "passage":       "Улуу Кыргыз кагандыгы -- 9-кылымда Енисей ...",
  "question":      "Улуу Кыргыз кагандыгы кайсы кылымда күчөп турган?",
  "choices":       ["9-кылымда.", "8-кылымда.",
                    "10-кылымда.", "7-кылымда."],
  "answer_index":  0
}
\end{verbatim}
\end{document}